\documentclass[conference]{IEEEtran}
\IEEEoverridecommandlockouts

\usepackage{amsmath,amssymb,amsfonts}
\usepackage{graphicx}
\usepackage{xcolor}
\usepackage{makecell}
\usepackage{array}

\usepackage[disable]{todonotes}
\usepackage{enumitem}
\usepackage{hyperref}
\usepackage{cleveref}
\usepackage{multicol}
\usepackage{subcaption}
\usepackage{mathtools}
\usepackage{tabularx}
\usepackage{booktabs}
\usepackage{clipboard}

\newcommand{\mypar}[1]{\smallskip\noindent\textbf{#1.}}

\newtheorem{axiom}{Axiom}

\definecolor{rebuttal}{rgb}{0.0, 0.0, 0.0}
\definecolor{minor}{rgb}{0.0, 0.0, 0.0}

\definecolor{needsrework}{rgb}{0.9, 0.1,0.0}

\definecolor{brainstorm}{rgb}{0.0,0.6,0.6}
\definecolor{first}{rgb}{0.85,0.5,0.5}
\definecolor{ok}{rgb}{0.4,0.3,0.3}
\definecolor{happy}{rgb}{0,0.0,0.0}

\def\ourmethod{SHAining }

\IEEEoverridecommandlockouts
\IEEEpubid{\parbox{\textwidth}{\normalfont\footnotesize%
\vspace{45pt}
\textbf{\copyright 2025 IEEE. Personal use of this material is permitted. Permission from IEEE must be obtained for all other uses, in any current or future media,}\\
\textbf{including reprinting/republishing this material for advertising or promotional purposes, creating new collective works, for resale or redistribution to servers or lists, or reuse of any copyrighted component of this work in other works.}
\hfill}
\hspace{\columnsep}\hfil}

\begin{document}

\title{SHAining on Process Mining: Explaining Event Log Characteristics Impact on Algorithms}

\author{
\IEEEauthorblockN{
Andrea Maldonado\IEEEauthorrefmark{1}\IEEEauthorrefmark{2}\IEEEauthorrefmark{3}, 
Christian M. M. Frey\IEEEauthorrefmark{6},
Sai Anirudh Aryasomayajula\IEEEauthorrefmark{1},\\ 
Ludwig Zellner\IEEEauthorrefmark{1}, 
Stephan A. Fahrenkrog-Petersen\IEEEauthorrefmark{4}, 
Thomas Seidl\IEEEauthorrefmark{1}\IEEEauthorrefmark{2} 
}
\IEEEauthorblockA{\IEEEauthorrefmark{1}Database Systems and Data Mining, Ludwig Maximilian University of Munich, Germany\\
\IEEEauthorrefmark{2}Munich Center for Machine Learning, Germany\\
\IEEEauthorrefmark{3}School of Engineering and Design, Technical University of Munich, Germany\\
\IEEEauthorrefmark{6}Machine Learning Lab, University of Technology Nuremberg, Germany\\
\IEEEauthorrefmark{4}University of Liechtenstein, Liechtenstein\\
andrea.maldonado@tum.de,
christian.frey@utn.de, 
anirudhsai027@gmail.com,\\
zellner@dbs.ifi.lmu.de, 
stephan.fahrenkrog@uni.li,
seidl@dbs.ifi.lmu.de}
}
\maketitle

\begin{abstract}
\textcolor{minor}{Process mining}\todo{R1.6} aims to extract and analyze insights from event logs, yet algorithm \textcolor{rebuttal}{\todo{R1.4}metric results} vary widely depending on structural event log characteristics.
Existing work often evaluates algorithms on a fixed set of real-world event logs but lacks a systematic analysis of how event log characteristics impact algorithm\textcolor{rebuttal}{s} individually.
Moreover, since event logs are generated from processes, where characteristics co-occur, we focus on associational rather than causal effects to assess how strong the overlapping individual characteristic affects \textcolor{rebuttal}{\todo{R1.4}evaluation metrics} without assuming isolated causal effects, a factor often neglected by prior work.
We introduce SHAining, the first approach to quantify the marginal contribution of varying event log characteristics to process mining \textcolor{rebuttal}{algorithms' metrics}.
Using \textcolor{minor}{process discovery}\todo{R1.6}  as a downstream task, we analyze over 22,000 event logs covering a wide span of characteristics to uncover which affect \textcolor{rebuttal}{algorithms across metrics}  (e.g., fitness, precision, complexity) the most.
Furthermore, we offer novel insights about how the value of event log characteristics correlates with their contributed impact, assessing the algorithm’s robustness. 
\end{abstract}


\section{Introduction}
\label{sec:intro}

\textcolor{happy}{
Process mining (PM) techniques are widely used to extract actionable insights from event logs across various domains such as healthcare, manufacturing, and finance.
Among these techniques, process discovery, which constructs process models from event logs, has received substantial attention~\cite{rehse2024process, van2021all}.
However, evaluating the quality of discovered models remains a challenge, particularly due to the heterogeneity in event log characteristics~\cite{maldonado2024gedi, burattin2022purpose, jouck2019generating}.
Event logs can differ significantly in structural properties.
For instance, healthcare processes typically yield highly variable traces where most cases exhibit similar behavior~\cite{munoz2022process}, whereas logs from structured domains like manufacturing often show recurring patterns with a few dominant variants.}

\textcolor{happy}{
Yet, not all process discovery algorithms are equally equipped to handle variability in event log characteristics.
Previous studies have related algorithmic metrics to either structural descriptions of process models ~\cite{huang2024proreco, schreiber2021exploring} or event log characteristics ~\cite{vanden2014uncovering} via statistical approaches such as linear regression, assuming a fixed dataset and a fixed model.
Moreover, these approaches assume properties like homoscedasticity and feature independence, which are rarely satisfied in practice.
Prior work has examined how variation in event log feature values influences overall algorithm evaluation metrics; the question of \textit{``why''} a given technique performs well or poorly on a specific log remains largely unaddressed. 
\smallskip \\
}
\begin{figure}[t]
    \centering
    \includegraphics[width=\linewidth]{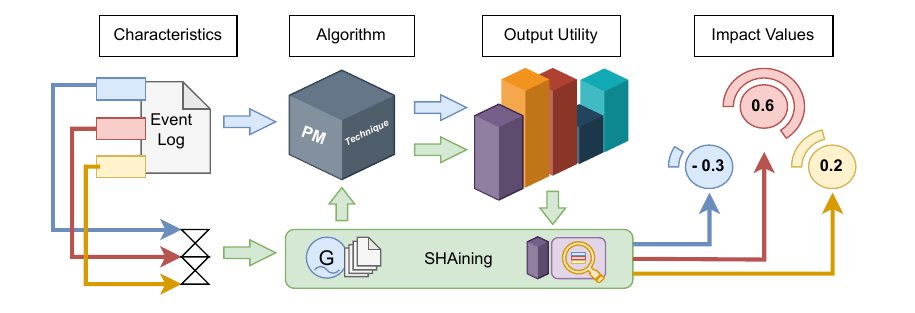}
    \caption{Studying the impact of log \textcolor{minor}{characteristics} on algorithm output with SHAining.}
    \label{fig:motivation}
    \vspace{-1.2em}
\end{figure}
\textcolor{happy}{
In this paper, we present the first systematic and explainable analysis of how overlapping event log characteristics individually contribute to impact onprocess mining algorithms.
Rather than optimizing algorithm selection for a downstream task, our goal is to explain metric variation by treating process mining techniques as black-boxes and analyzing how structural log characteristics affect evaluation outcomes.
To this end, we introduce SHAining,
depicted as the green bottom box in \Cref{fig:motivation}\todo{R3.5}, which quantifies the impact 
of characteristics, highlighted in \{blue, red, yellow\}, to algorithm metrics in a technique-agnostic way.
We generate over 22,000 intentional logs with varied structural properties using GEDI~\cite{maldonado2024gedi}, and evaluate multiple process mining algorithms with standard metrics such as fitness, precision, and F-score.
We apply a Shapley value analysis \cite{1952_shapley, Lipovetsky_2020_handbook} originating from game theory that fairly attributes the impact across all possible feature value combinations. We identify which feature values, e.g. entropy or variant diversity, affect an algorithm's performance the most.
In our analysis, we study the impact of event log characteristics on algorithm metrics using process discovery as a downstream task.
We demonstrate that contributions of event log characteristics to algorithm metrics vary by the type of structural characteristic and by their specific values.
For instance, high activity entropy degrades the output of some algorithms, revealing key trade-offs in algorithm suitability.
}

Overall, our key contributions are:
(1) Extends Shapley analysis to distributions defined by generative models conditioned on meta-feature configurations.
(2) Enables explainability of how distributional properties influence downstream model behavior.
(3) Large-scale analysis on 22,000+ synthetic event logs, revealing consistent feature-metric interactions.
%
%
\section{Related Work}
\label{sec:relatedWork}
Understanding the impact of event log characteristics on process discovery (PD) algorithms has become a growing area of interest in PM. A prime 
example is the Process Discovery Contest \cite{carmona2017summary} which annually compares PD algorithms 
to determine the best-performing approach.

\mypar{Explainability} Shapley value analysis is applied across various domains. Stevens et al.~\cite{stevens2021quantifying} apply Shapley value analysis to compare interpretable models and post-hoc explainability methods for predicting loan application outcomes. Similarly, Pishgar et al.~\cite{pishgar2022process} use Shapley values to quantify event attribute importance in predicting COVID-19 mortality. Heskes et al.~\cite{heskes2020causal} extend Shapley-based explanations by incorporating causal knowledge, highlighting the importance of distinguishing between association and causation when interpreting model outputs. Although these works demonstrate the usefulness of Shapley values for explainability in PM scenarios, they focus on predictive tasks rather than directly evaluating PD algorithms or explaining their behavior on specific log structures.
A closer line of research is the ProReco framework~\cite{huang2024proreco}, a recommender system that predicts performance metrics for different PD algorithms based on log characteristics and user preferences. While ProReco uses SHAP (\textbf{SH}apley \textbf{A}dditive ex\textbf{P}lanations) \cite{lundberg2017shap} to provide explanations, it applies SHAP on a surrogate model rather than the PD algorithms themselves. Additionally, it relies on heuristic imputation for incomplete data, and is limited to a much smaller synthetic dataset. 

\textcolor{rebuttal}{\todo{M2.1+\\R1.2\\+R2.3} A recent critique \cite{explainabilityNotAGame} argues that Shapley values can invalidate intuitions and induce false trust by assigning high scores to irrelevant features.
They propose applying formal methods, such as abductive/contrastive explanations (AXps/CXps) \cite{2024_logicBasedXAI}.
However, this critique targets the usage of Shapley analysis on a fixed per-instance level, i.e., explaining individual predictions. 
DShapley \cite{kwon_dshapley_2021} extends the classical Shapley value to quantify the contribution of features w.r.t. the expected utility over an underlying data distribution, rather than a fixed dataset. 
In our work, we compute Shapley values on meta-features that govern data generation to analyze impacts on distributions (cf. \cite{kwon_dshapley_2021}) of algorithm performance, not on individual predictions of a \emph{fixed} model on a \emph{fixed} input.
Hence, \ourmethod differs fundamentally from the fixed data scenario that AXps/CXps address.
Therefore, Shapley values over distributions \cite{kwon_dshapley_2021} arise as the most suitable for our analysis of associational trends across generative configurations.
}


\mypar{Benchmark studies} Several benchmark studies compare PD algorithms across diverse logs and quality metrics. Augusto et al.~\cite{augusto2018automated} evaluate multiple algorithms across 24 real-world logs using nine different evaluation criteria, providing insights into the strengths and weaknesses of these algorithms. Van den Broucke et al. \cite{vanden2014uncovering} analyze how structural log characteristics influence PD outcomes using metrics such as fitness and precision. Yet, their reliance on linear regression introduces strong assumptions such as feature independence and homoscedasticity that are often breached in practice due to the complex dependencies and variance inherent in real-life event logs.
A complementary direction is taken by Andree et al.~\cite{andree2024workflow}, who compare PD algorithms based on their ability to reproduce control-flow patterns. However, their focus is on what patterns are captured, rather than how structural log characteristics influence evaluation outcomes, a gap our work addresses directly.

\mypar{Generation} Recent work for data generation has tackled the scarcity and imbalance of benchmark logs. Maldonado et al. ~\cite{maldonado2024gedi} introduce GEDI, a framework for generating logs with diverse structural characteristics, enabling controlled experimentation. Similarly, Jouck et al. \cite{jouck2019generating} propose a generator based on random sampling from predefined process populations. Janssenwillen et al. \cite{janssenswillen2017comparative} explore how noise injection affects fitness and precision in PD evaluations, particularly for bias estimation. While these efforts improve empirical validity, they stop short of explicitly analyzing how individual log characteristics affect evaluation outcomes.
%
%
%
\section{Preliminaries}
\label{sec:preliminaries}
\mypar{Event Log}
An event $e \in \mathcal{E}$ represents a step in a process and is characterized by $e:=(c,a,t)$, with a case identifier $c \in \mathcal{C}$, an activity $a \in \mathcal{A}$
describing the type of event, and a timestamp $t \in \mathcal{T}$
when the event occurred.
A sequence of events is called a trace $\sigma := \langle e_1, e_2, \ldots, e_n \rangle$ and groups events that belong to the same case within the process. Within a trace, events are ordered based on their timestamp. Finally, we define a multiset of traces as an event log $L$. We denote the universe of all possible event logs as $\mathcal{L}$.

\mypar{Event Log characteristics}
\textcolor{happy}{Event logs capture the execution of processes, which can exhibit a wide spectrum of characteristics, ranging from highly structured to highly variable and complex workflows.}
In our work, we quantify these characteristics using \emph{event log features} \( \mathcal{F} := \{F_1, \dots, F_n\} \),
which describe different aspects related to traces, variants, activities, and events, such as trace lengths, and activity frequencies. 
Following Maldonado et al.~\cite{maldonado23feeed}, feature extraction is formalized as a function \( f_e: \mathcal{L} \rightarrow \mathbb{R}^n \), mapping an event log \( L \in \mathcal{L} \) to an \( n \)-dimensional feature vector \( f := (f_1, \dots, f_n) \), where each \( f_i \in \mathbb{R} \) represents a specific feature value.

\mypar{Shapley Value Analysis}
Shapley value analysis \cite{1952_shapley} originates from game theory, where we define a set of players $N=\{1, \ldots, n\}$ and a characteristic function $v:2^N \rightarrow \mathbb{R}$ that assigns to each coalition of players $S \subseteq N$ a real number $v(S)$\footnote{The condition $v(\emptyset)=0$ holds per definition of Shapley values.}. 
This concept has been widely adapted in machine learning to interpret model predictions, where "players" correspond to input features, and $v(S)$ represents the model’s output when only the subset $S$ of features is considered. The Shapley value thus offers a rigorous, fair method to evaluate the influence of each feature on a model's prediction.
\textcolor{rebuttal}{
In our work, we examine the impacts of meta-features on the data-generating process, rather than on fixed input datasets.
}

\section{SHAining - Shapley values for relating process mining tasks and event log features}%
\label{sec:methodology}%
\begin{figure*}[!ht]
    \centering\includegraphics[width=\linewidth]{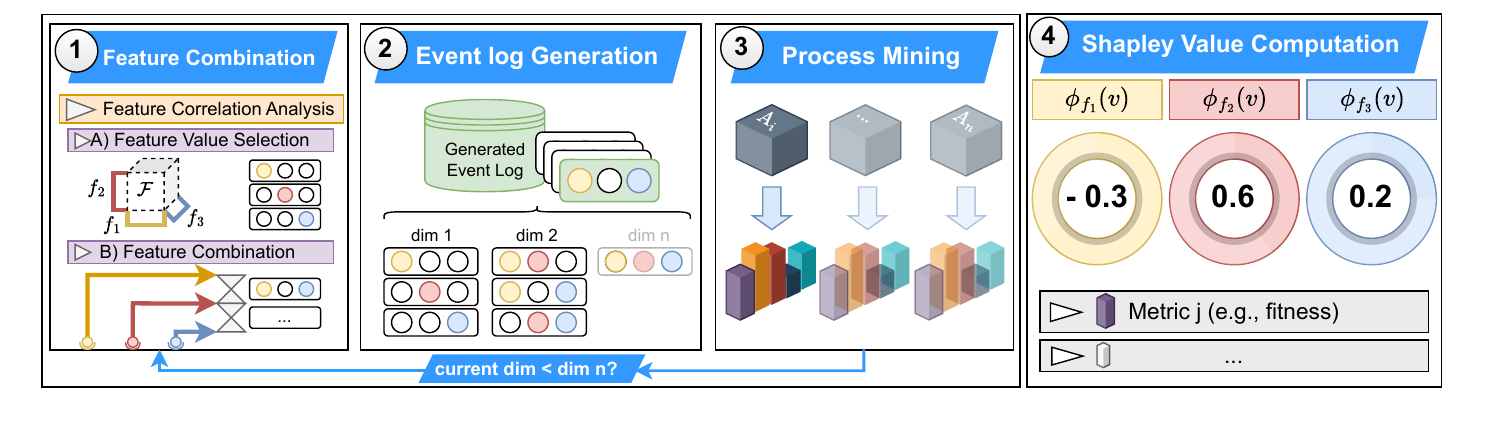}
    \vspace{-3em}
    \caption{The workflow of SHAining.}
    \label{fig:workflow}
    \vspace{-1.2em}
\end{figure*}

\textcolor{happy}{
We apply Shapley values \cite{1952_shapley,Lipovetsky_2020_handbook,lundberg2017shap} for a higher-order problem 
across the space of data-generating processes. 
Concretely, our players represent 
features in the data generation process itself. 
For a well-defined characteristic function, we apply a Shapley analysis, where the \emph{players} are (meta-)features that generate logs, and the \emph{game} is defined by a model's performance. 
%
%
More concretely, we answer the question: \emph{What is the marginal value of including feature $F_i$ in the log generation process, in terms of the downstream utility of a PM technique?}
Therefore, we are not comparing feature subsets for a \emph{fixed} model on a \emph{fixed} dataset, but rather assessing the marginal effect of a feature on the generative process, and through it, on a model's evaluation results.
}

\textcolor{happy}{
Our approach consists of four steps: 
\textbf{(i) Feature Combination.~}
\textbf{(ii) Event log Generation.~}
\textbf{(iii) Process Mining.~}
\textbf{(iv) Shapley Value Computation.~}
The first three steps are repeatedly applied for a robust Shapley Value analysis across a diverse set of event logs being processed by a black-box model.~
An overview of \ourmethod is summarized in \autoref{fig:workflow}.}
In this section, we elaborate on each step individually:

\mypar{(i) Feature Combination}
As in \cref{sec:preliminaries}, let the domain of (meta-)features be $\mathcal{F}$. 
We define an n-dimensional configuration vector as a specific realization of features: 
\todo{R3.4+\\R.2.2}
\textcolor{rebuttal}{
A) First, we initialize 1-dimensional vectors, each as $f_{S_i}\coloneqq \{(F_i=f_{S_i}) | F_i \in \mathcal{F}\}$, each containing a value for the $i$-th feature; 
B) Later, any $(|S_i|+1)$-dimensional vector is $f_{S_i \oplus S_j} \coloneqq f_{S_i} \oplus f_{S_j}$ composed of two $|S_i|$-dimensional feature vectors 
$f_{S_i},f_{S_j} \in \mathbb{R}^{|S|}$, of a subset $S \subseteq \mathcal{F}$ of features 
, where $\oplus$ is the subspace join operator, resulting in the smallest common superspace.} 
%
In our evaluation, we utilize a correlation analysis on features to define $\mathcal{F}$ (see \autoref{sec:evaluation}).

\mypar{(ii) Event log Generation} We leverage an event log generator to produce different event logs \textcolor{rebuttal}{\todo{R2.2}targeting respective feature configurations. }%
Generally, assuming that we have access to a prior over all event logs in the universe, denoted as $L \sim p(L)$, from which we can sample infinitely many logs. 
In that case, we can analyze an infinite number of feature combinations and their resulting metric distributions. 
In our work, we assume $p(L | f_S)$ to be a parameterized prior distribution over event logs with a feature configuration vector $f_S$ controlling the feature distributions.
\todo{R2.3} The log generator $\mathcal{G}$ is defined as a sampling function $\mathcal{G}: f_{S} \mapsto L \sim p(L | f_{S})$. 
As a generator, we leverage GEDI~\cite{maldonado2024gedi} as prior, resulting in event logs \textcolor{rebuttal}{that align features to \todo{R2.2}target feature values in the configuration vector}.

\mypar{(iii) Process Mining}
We leverage a utility function \textcolor{rebuttal}{$U(A, L_{S})$} to measure a black-box model's quality in terms of evaluation metrics, when applying a fixed process mining algorithm $A$ on the data \textcolor{rebuttal}{$L_{S}$}. We define: 
\begin{equation}%
\label{eq:utility}
v(S) \coloneqq U(A, L_{S})
\end{equation}
In our evaluation, we apply \textcolor{rebuttal} {\Cref{eq:utility}} 
for $\mathcal{A}$ different black box models and $\mathcal{M}$ metrics \textcolor{rebuttal}{on \todo{R2.2}each generated event log.
Whenever a feature configuration could be generated, and the log passed the process mining analysis (steps i-iii), we evaluate whether the dimension of the configuration feature vector equals the number of players. 
If it does not, we (i) combine the current feature vectors pair-wise, otherwise we continue to (iv) Shapley Value Computation.}

\mypar{(iv) Shapley Value Computation}
We define the functional Shapley value for a feature $f_i \in \mathcal{F}$  as \cite{Lipovetsky_2020_handbook}, \textcolor{rebuttal}{\todo{R2.2}using the PM evaluation measurements from utility function $v(S)$.
By averaging the marginal contributions across all coalitions, we get the final Shapley value per feature that quantifies its average impact on an algorithm's evaluation results:}
\begin{small}
\begin{equation}
\label{eq:shapley}
\phi_i =\sum_{S \subseteq \mathcal{F} \setminus \{f_i\}}\frac{|S|!(|\mathcal{F}|-|S|-1)!}{|\mathcal{F}|}[v(S \cup \{f_i\}) - v(S)],
\end{equation}%
\end{small}%
\noindent%
\mypar{Discussion}
The following properties are fullfilled\cite{Lipovetsky_2020_handbook}:
{
\begin{axiom}[Efficiency]%
\label{ax:efficiency}
The total gain of the feature set $\mathcal{M}$ is distributed:
$\sum_{i=0}^{|\mathcal{F}|} \phi_i = v(\mathcal{F})$
\end{axiom}
\begin{axiom}[Symmetry]%
\label{ax:symmetry}
If two features $f_i, f_j \in \mathcal{F}$ contribute equally to all possible coalitions, i.e., $v(S \cup \{f_i\}) = v(S \cup \{f_j\})$, then their Shapley values are identical: 
$\phi_i = \phi_j$
\end{axiom}
\begin{axiom}[Additivity]%
\label{ax:additivity}
If two coalitions of features defined by gain functions $v$ and $w$ are combined, then the distributed gains correspond to gains derived from $v$ and $w$, i.e.,
$\phi_i (v + w) = \phi(v) + \phi(w)$
Additionally, for any scalar $a \in \mathbb{R}$, we have $\phi(a\cdot v) = a \phi_i(v)$.
\end{axiom}
\begin{axiom}[Null player]
\label{ax:null}
If it holds that $v(S \cup f_i) = v(S), \forall S \subseteq \mathcal{F} \setminus \{f_i\}$, then the feature $m_i$ is called a null player with $\phi_i=0$.
\end{axiom}}

Axioms \textbf{Additivity} \ref{ax:additivity}  and \textbf{Null Player} \ref{ax:null} 
are trivially satisfied: the former holds as a null player's marginal contributions to any coalition are zero;
the latter directly follows from linearity in the second term.
Our application satisfies the \textbf{Symmetry} \ref{ax:symmetry} axiom as our utility function (cf. \Cref{eq:utility}) does not distinguish between features that induce identical shifts in data distributions and lead to identical model performance.
In such cases, their marginal contributions are the same across all permutations.
Finally, despite our utility function $v(S)$ being based on a fixed model applied to \textcolor{minor}{non-fixed} data sampled from a conditioned generative process, the \textbf{Efficiency} \ref{ax:efficiency} axiom still holds \textcolor{minor}{due to}
the combinatorial structure used to compute each feature's marginal contribution across all permutations, as shown by rewriting \Cref{eq:shapley} as:
\textcolor{happy}{
\vspace{-0.5em}
\begin{align}
\sum_{i=0}^{|\mathcal{F}|}\phi_i &= \frac{1}{|\mathcal{F}|!} \sum_{\pi \in Perm(\mathcal{F})} \sum_{j=1}^{|\mathcal{F}|}\left[ v(P_{\pi}(f_j) \cup \{f_j\} - v(P_{\pi}(f_j)))\right]\nonumber\\
&=\frac{1}{|\mathcal{F}|!} \sum_{\pi \in Perm(\mathcal{F})}\left[ v(\mathcal{F}) - v(\emptyset) \right] = v(\mathcal{F})\nonumber,
\end{align}
}
\noindent%
where $P_{\pi}$ denotes the set of players \textbf{p}receding the addition of feature $f_i$. 
When aggregating these contributions, the marginal effects follow a telescoping sum across all permutations spanning the entire value of $v(\mathcal{F})$. In our definition, even though datasets are different for each coalition, the utility function is a well-defined set function over $2^{\mathcal{F}} \rightarrow \mathbb{R}$ yielding a scalar value for all coalitions. Thus, even \textcolor{minor}{though $v(S)$ is induced}\todo{R1.6} by a different data distribution, the Shapley value ensures that efficiency is preserved in our setting.

\section{SHAining on Process Discovery}
\label{sec:evaluation}

\indent 
\textcolor{rebuttal}{We answer the following research questions:}
\begin{description}
   \item[RQ1] \Copy{RQ1}{\textcolor{minor}{\todo{R1.6}Can we reveal which EL characteristics contribute the most to the algorithm's} \textcolor{rebuttal}{\todo{R1.4}evaluation results}\textcolor{minor}{?}}
    \item[RQ2] \Copy{RQ2}{\textcolor{minor}{\todo{R1.6}Do EL characteristics' impacts on algorithm \textcolor{rebuttal}{\todo{R1.4}evaluation results }\textcolor{minor}{change depending on feature values?}}}
    \item[RQ3] \Copy{RQ3}{How can insights about the impact of EL characteristics support \textcolor{rebuttal}{\todo{M3\\+R1.4}process miners} on a specific downstream task?}
\end{description}


\mypar{Setup and Implementation Details}
\textcolor{happy}{We run our framework on a
Intel(R) Xeon(R) Platinum 8160 CPU @ 2.10GHz using 239Gi RAM.
Our code is publicly available\footnote{\label{foot:rep}\href{https://github.com/andreamalhera/SHAining/tree/icpm25}{https://github.com/andreamalhera/SHAining/tree/icpm25}}.
Furthermore, each algorithm was tested under resource constraints with a maximum of 5 minutes and 19GB of disk storage per log.
}

\mypar{Features}
%
For our experiments, we selected a subset from FEEED \cite{maldonado23feeed} using a greedy approach.
This method minimizes inter-feature correlation while maintaining representative coverage.
As in Huang et al. \cite{huang2024proreco}, we use the Pearson correlation coefficient to identify representative meta-features from different groups, e.g., activity-/time-/trace-based features.
We iteratively chose the meta-feature with the lowest average correlation and then added the feature with the smallest maximum correlation to the selected set.
Using the elbow method, we set a cutoff point of $n$=8 meta-features.
More details on this pre-selection step are available in our repository.
Moreover, we select 
the features with ten equidistant value samples.
Thus, the feature set in our evaluation is \(F = \{\text{aq1},  \text{nusa}, \text{saq1}, \text{ekbr3}, \text{rt5v}, \text{svo}, \text{tlkh}, \text{tlv}\}\), with:
\textcolor{happy}{
\begin{description}
\item[activities\_q1 (aq1):] Lowest 25\% (quartile) of activity counts in the log. Range: [1.0, 79.92]. 
\item[n\_unique\_start\_activities (nusa):] Counts unique start activities in the log, indicating process diversity at the start. Range: [1.0, 6.56].    
\item[start\_activities\_q1 (saq1):] Lowest 25\% of start activity counts in the log. Range: [1.0, 174.79].
\item[eventropy\_k\_block\_ratio\_3 (ekbr3):] Normalized ratio of the 3-subsequence entropy in a log \cite{6138a20588854e0182f4aa3595788ac6}. Range: [0.0, 4.37].
\item[ratio\_top\_5\_variants (rt5v):]  The proportion of traces in the top 5\% most frequent variants. Range: [0.0, 0.38].
\item[skewness\_variant\_occurance (svo):]  Measures how unevenly process variants occur, showing if the distribution is balanced or not. Range: [1.54, 11.61].
\item[trace\_len\_kurtosis (tlkh):] Measures how much the trace lengths vary, indicating the concentration of items at the center. Range: [-0.97, 7.92].
\item[trace\_len\_variance (tlv):] Measures how much the lengths of different process variants vary. Range: [0.0, 138.7]. 
\end{description}
}
\mypar{Datasets} Logs are generated during SHAining's generation step (see \autoref{sec:methodology}), with the number of logs resulting to \(\sum_{k=1}^{\text{k}_{max}} \binom{n}{k} v^k\) where \(n\) is the number of features, \(v\) is the number of values per feature, and \(\text{k}_{max}\leq n\) is \textcolor{rebuttal}{\todo{R.1.6}the maximal number of features in one single configuration vector  (cf. \Cref{sec:methodology})}. 
In our evaluation, we set $\text{k}_{max}$=3 to three features. 
With $v$=10 values and $n$=8 features, the number of possible feature combinations yields 58,880 logs.
We refer to our repository for a summary of the event log statistics.

\mypar{PM Downstream Task}
\ourmethod's modular architecture, decoupled from black-box mechanics, enables seamless extension to various PM tasks.
To quantify Shapley Value insights in PM, we use \textcolor{minor}{\emph{process discovery (PD)}\todo{R1.6}} as a representative downstream task.
PD is a fundamental PM task that automatically generates process models from event logs.
Our evaluation compares three well-established \textcolor{minor}{\emph{PD algorithms}\todo{R1.6}}, selected from \textcolor{rebuttal}{different discovery paradigms \todo{R2.3}, as top-down vs. bottom-up,} for thorough assessment.
We use default hyperparameters from their original works.
Inductive Miner (\textit{IND})~\cite{leemans2013discovering} uses a top-down approach, recursively partitioning event logs to construct models.
Integer Linear Programming Miner (\textit{ILP})~\cite{van2018discovering} uses optimization to incrementally identify patterns bottom-up.
Split Miner (\textit{SPM})~\cite{augusto2019split} builds sound process models by examining directly-follows graphs and loops first.

For fair and robust analysis, we evaluate PD outputs using multiple \emph{\textcolor{minor}{metrics}\todo{R1.6}}, as in \cite{maldonado2024gedi} and \cite{augusto2022connection}:
\textcolor{minor}{
\begin{description}
    \item[Quality metrics:] \textit{Fitness} (Ft), \textit{precision} (Pr), \textit{F-score} (Fs)~\cite{6306ea88338545bc9f1781894c6736ef}; higher values indicate better model-log alignment~\cite{adriansyah2015measuring,adriansyah2011conformance}. 
    \item[Complexity metrics:]\textit{Size} (Sz), i.e. number of BPMN nodes, and \textit{control-flow complexity} (Cf), i.e. degree of split gateway branching~\cite{mendling2008metrics}.
    \item[Performance metrics:] \textit{Execution time} (Et) and model soundness, i.e., behavioral correctness \cite{van1997verification,augusto2018automated}.
\end{description}
} 

All metric evaluation results and Shapley Values per feature-value combination are in our repository.
Our remaining analysis focuses on the resulting Shapley Values.
For example, a Shapley value of 3.4 for trace length kurtosis (\textit{tlkh}) concerning \textit{Sz} and \textit{ILP} indicates this feature increases model size by 3.4 on average.
Normalized, a 0.22 value means \textit{tlkh} accounts for 22\% of the total feature impact on this metric.

\subsection{RQ1: \Paste{RQ1}}
\label{subsec:exp:rq1}
\begin{figure*}[t]
\centering
\begin{subfigure}{0.32\textwidth}
    \includegraphics[width=\textwidth, trim={1.25cm 0.6cm 1.25cm 0.8cm}, clip]{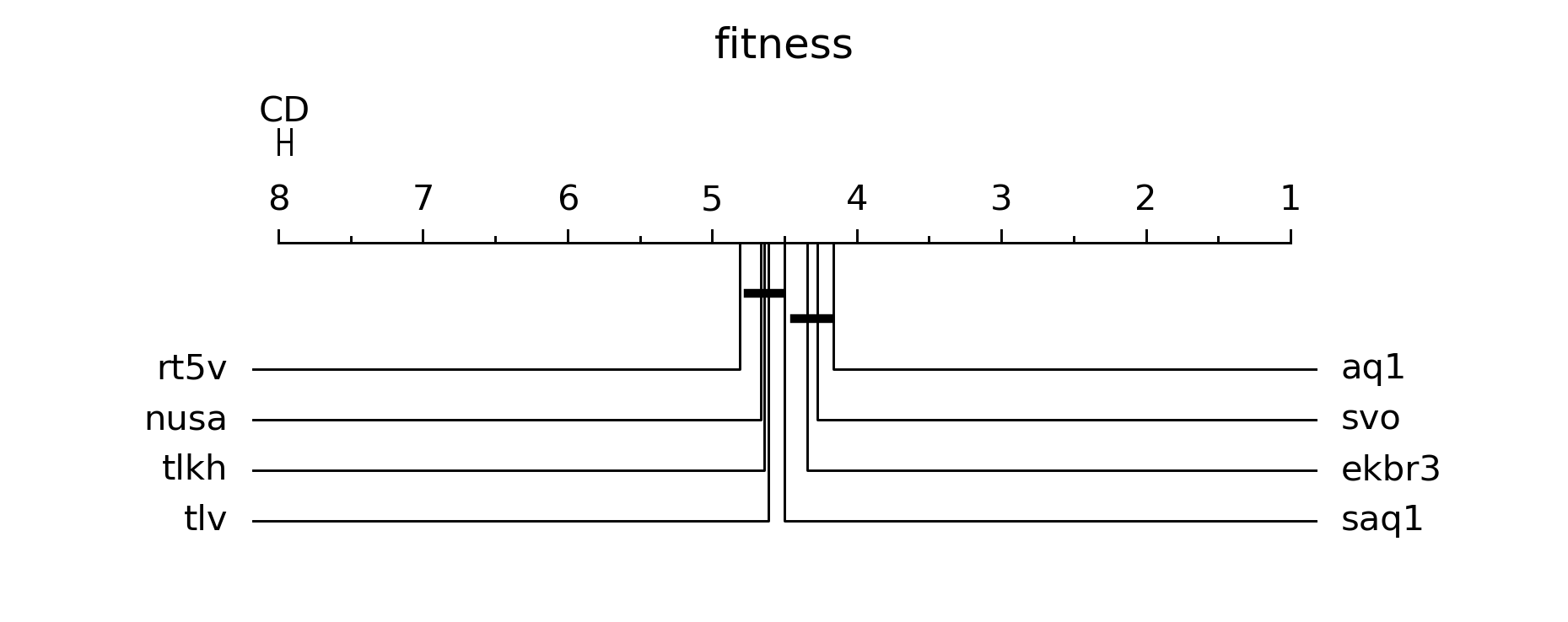}
    \caption{Fitness (Ft)}
    \label{fig:cd_fitness}
\end{subfigure}
\hfill
\begin{subfigure}{0.32\textwidth}
    \includegraphics[width=\textwidth, trim={1.25cm 0.6cm 1.25cm 0.8cm}, clip]{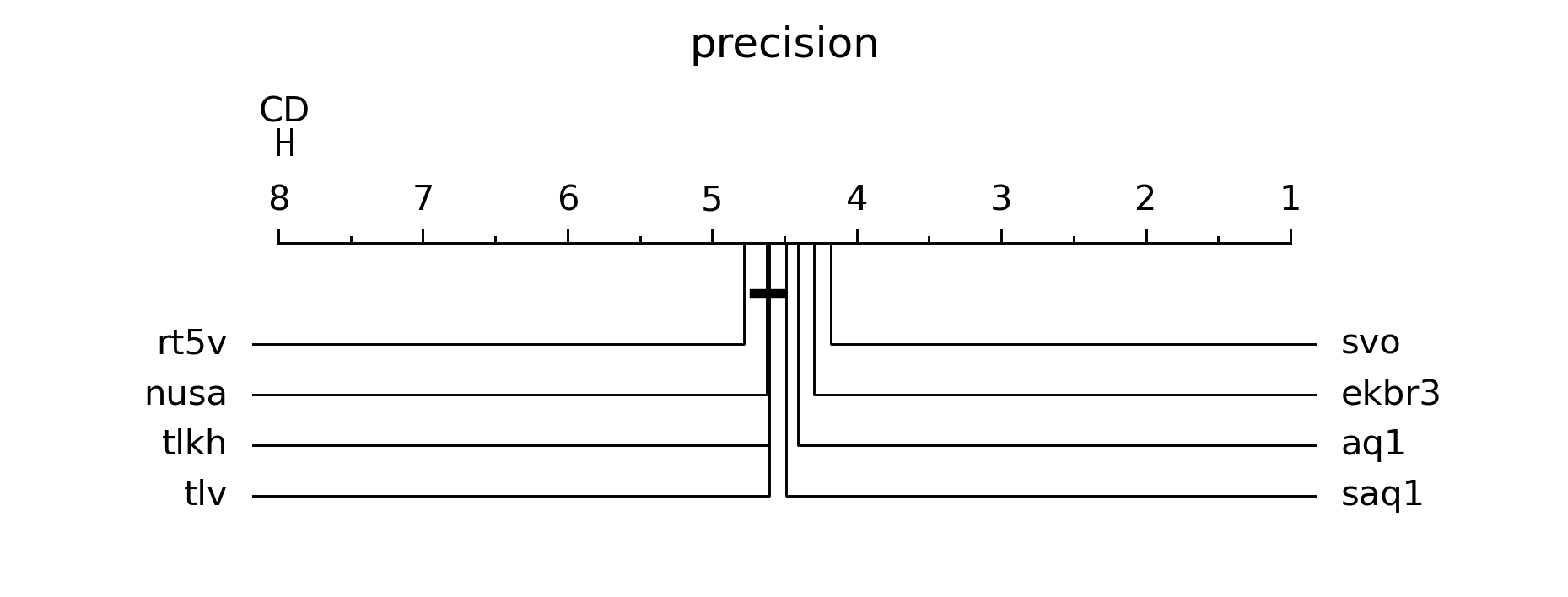}
    \caption{Precision (Pr)}
    \label{fig:cd_precision}
\end{subfigure}
\hfill
\begin{subfigure}{0.32\textwidth}
    \includegraphics[width=\textwidth, trim={1.25cm 0.6cm 1.25cm 0.8cm}, clip]{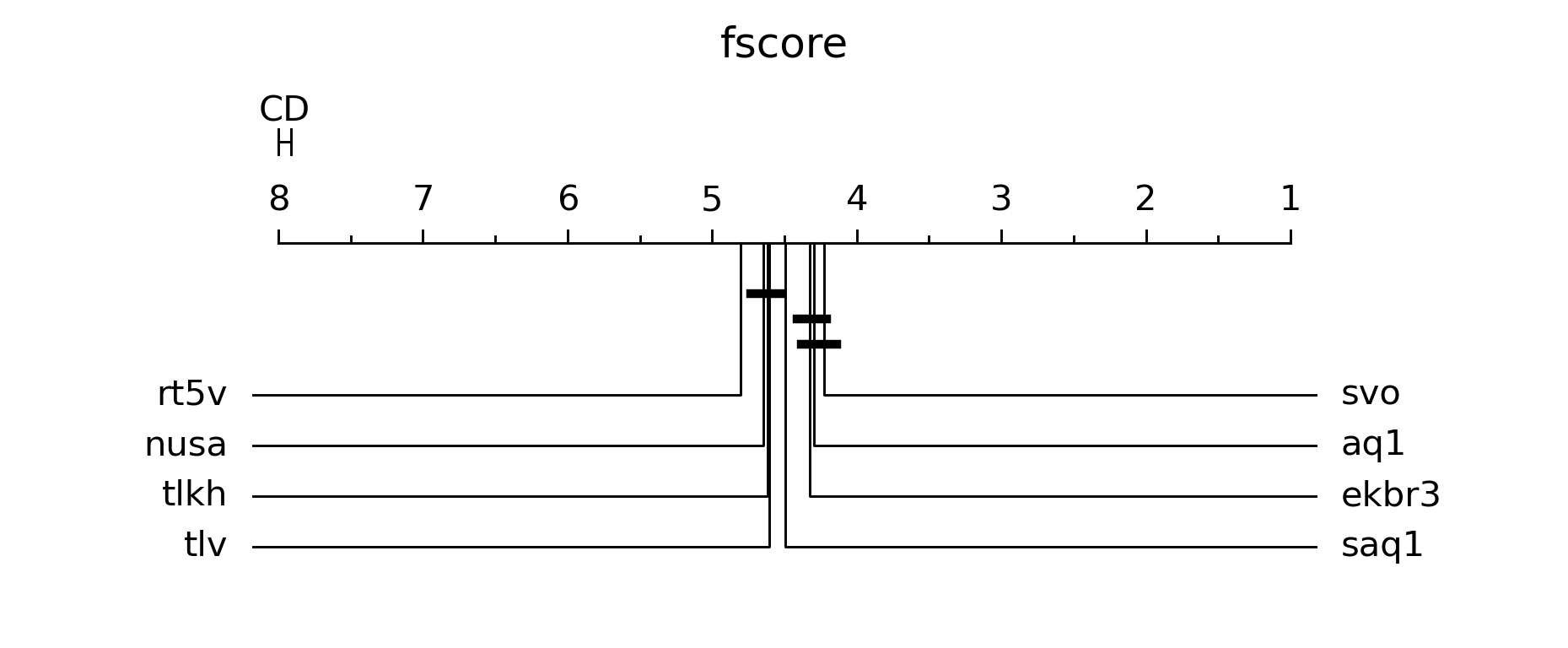}
    \caption{F-score (Fs)}
    \label{fig:cd_fscore}
\end{subfigure}
\begin{subfigure}{0.32\textwidth}
    \includegraphics[width=\textwidth, trim={1.25cm 0.6cm 1.25cm 0.8cm}, clip]{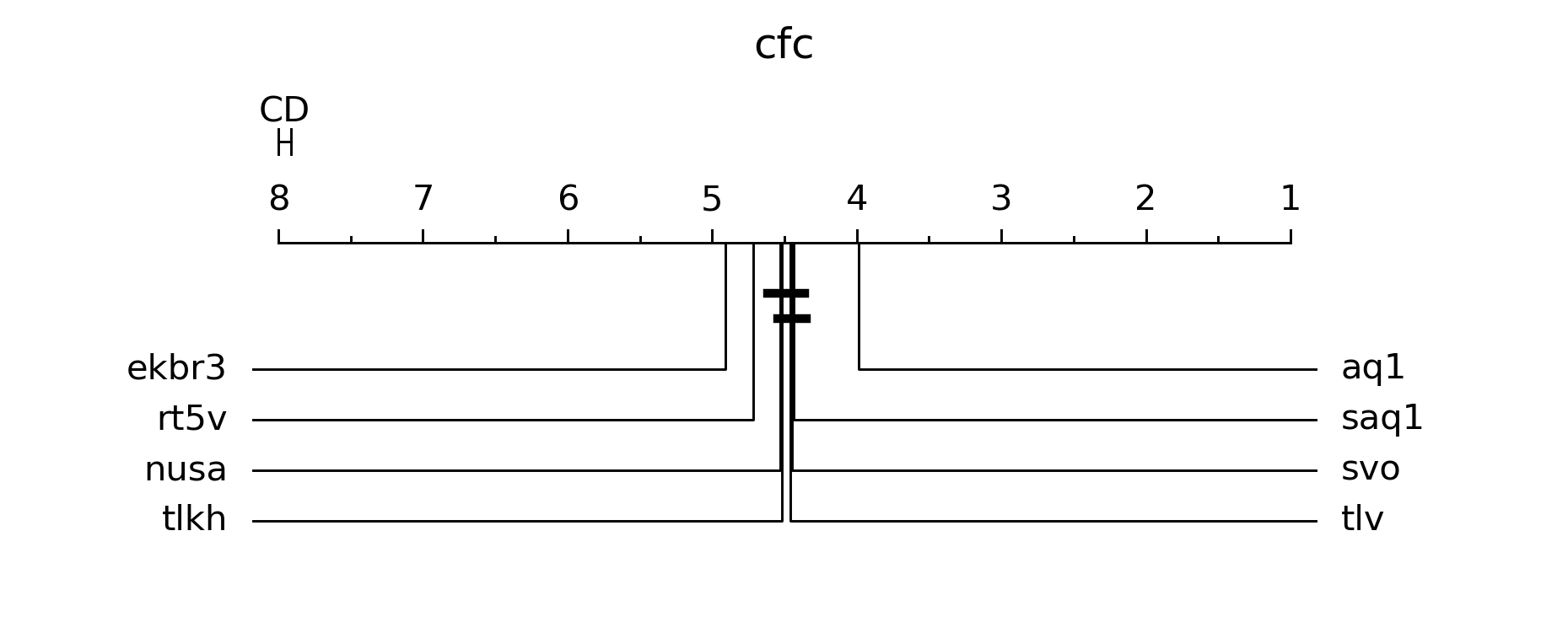}
    \caption{Control-flow-complexity (Cf)}
    \label{fig:cd_cfc}
\end{subfigure}
\hfill
\begin{subfigure}{0.32\textwidth}
    \includegraphics[width=\textwidth, trim={1.25cm 0.6cm 1.25cm 0.8cm}, clip]{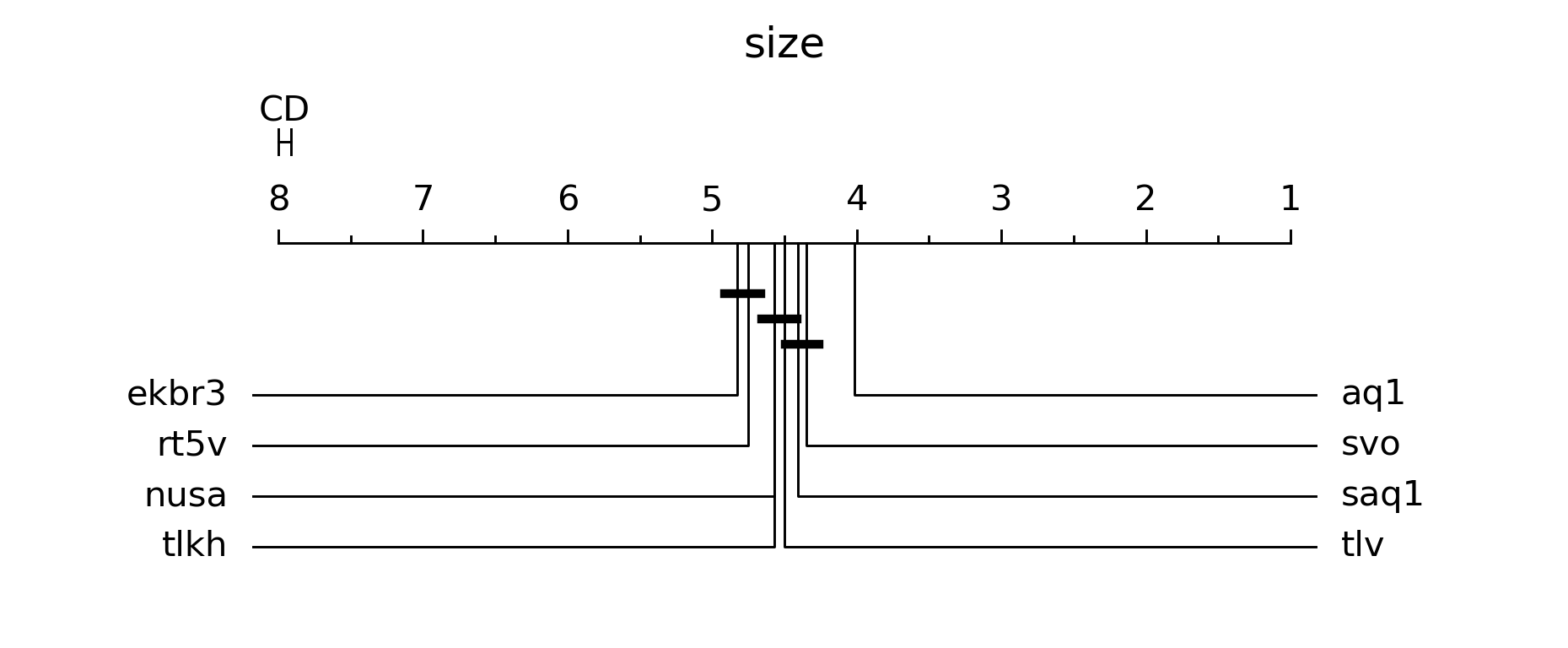}
    \caption{Size (Sz)}
    \label{fig:cd_size}
\end{subfigure}
\hfill
\begin{subfigure}{0.32\textwidth}
    \includegraphics[width=\textwidth, trim={1.25cm 0.6cm 1.25cm 0.8cm}, clip]{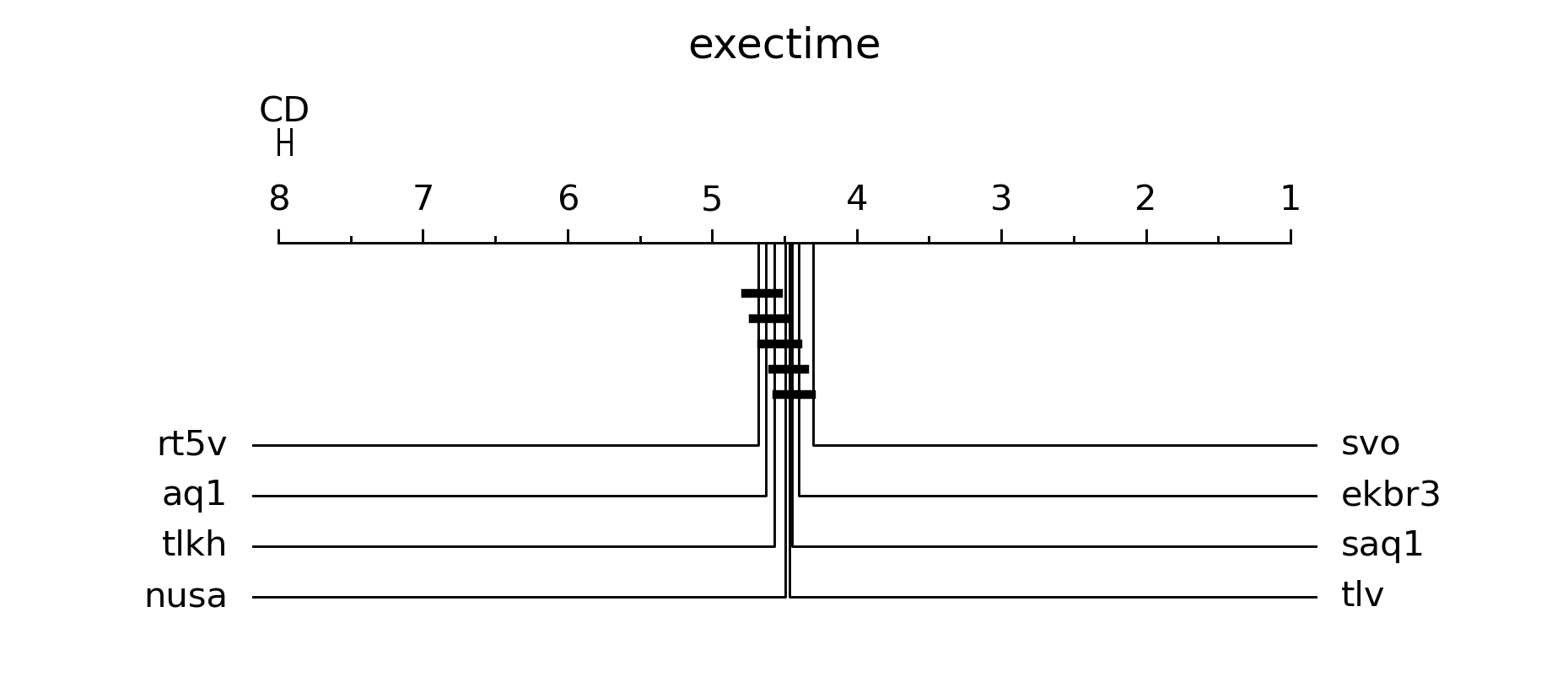}
    \caption{Exec. Time (Et)}
    \label{fig:cd_exectime}
\end{subfigure}
\caption{CD diagrams showing meta-features' contributions to three selected metrics across all evaluated PD algorithms.}
\label{fig:cd_mfs}
\vspace{-0.6em}
\end{figure*}

\begin{figure*}[t]
\centering
\begin{subfigure}{0.32\textwidth}
    \includegraphics[width=\textwidth, trim={1.25cm 0.6cm 1.25cm 0.8cm}, clip]{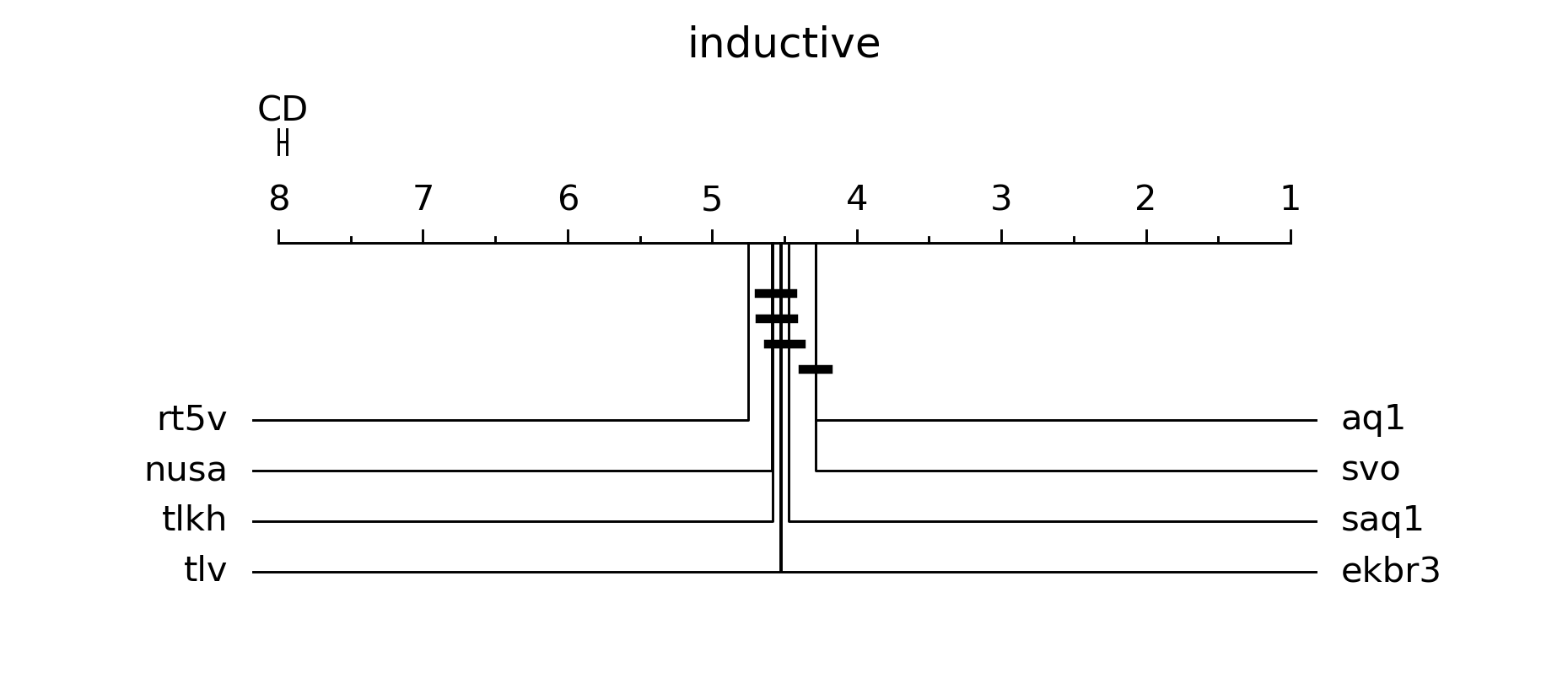}
    \caption{IND}
    \label{fig:cd_ind}
\end{subfigure}
\hfill
\begin{subfigure}{0.32\textwidth}
    \includegraphics[width=\textwidth, trim={1.25cm 0.6cm 1.25cm 0.8cm}, clip]{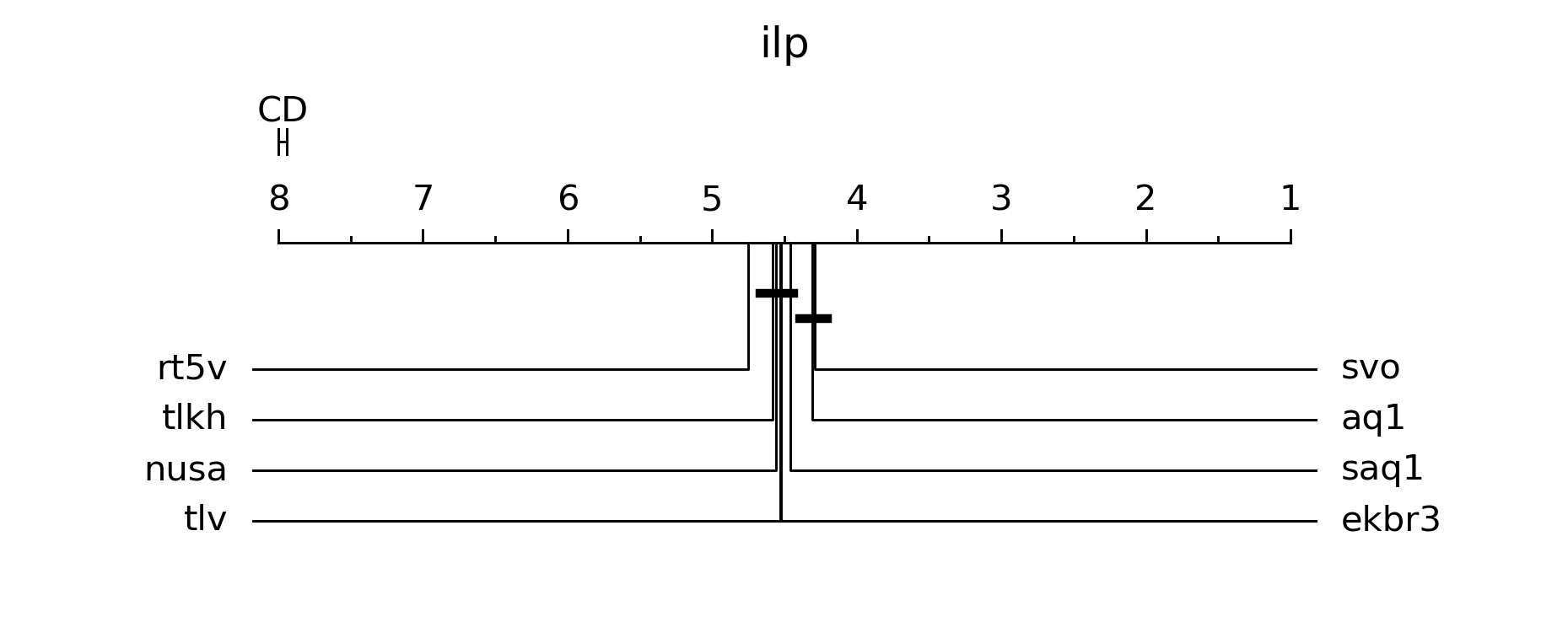}
    \caption{ILP}
    \label{fig:cd_ilp}
\end{subfigure}
\hfill
\begin{subfigure}{0.32\textwidth}
    \includegraphics[width=\textwidth, trim={1.25cm 0.6cm 1.25cm 0.8cm}, clip]{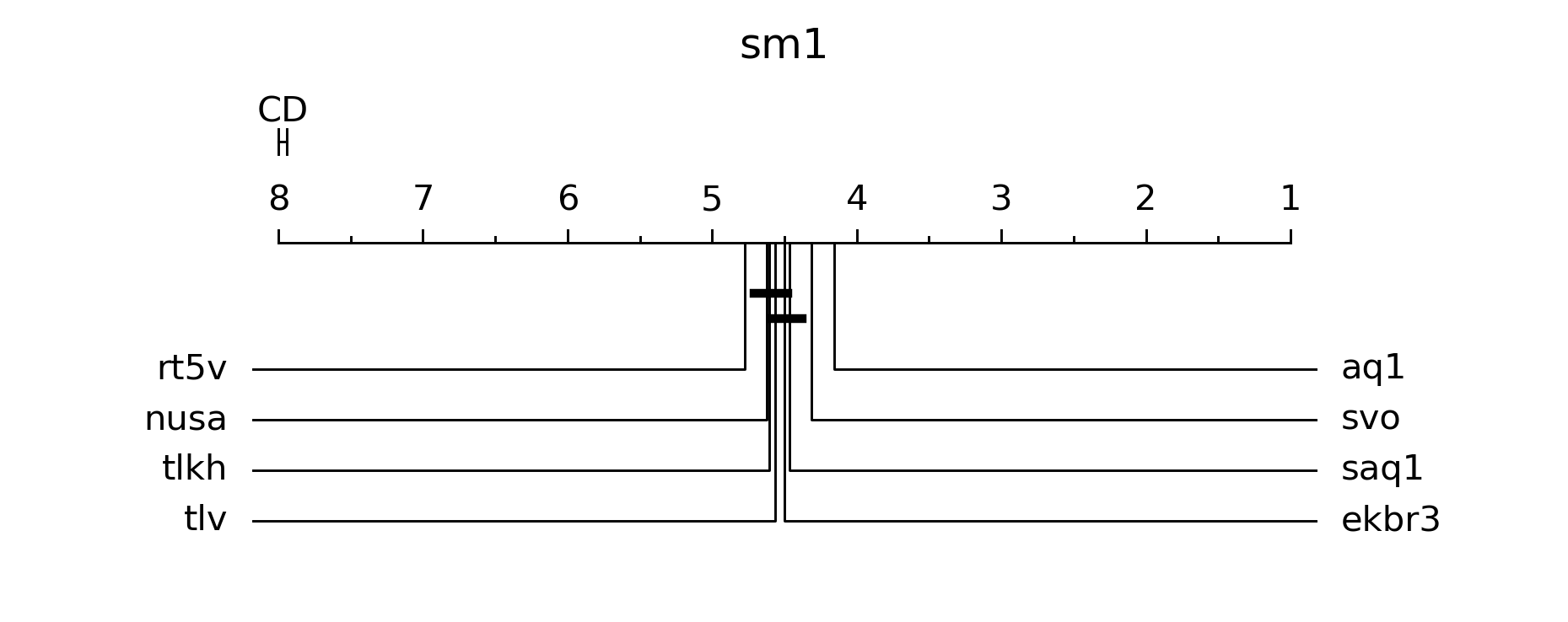}
    \caption{SPM}
    \label{fig:cd_sm}
\end{subfigure}
\caption{CD diagrams showing significant meta-features' contributions per PD algorithm across all metrics}
\label{fig:cd_miners}
\vspace{-0.6em}
\end{figure*}

\textcolor{rebuttal}{We\todo{M2.2+\\R1.3} evaluate the empirical average marginal effect over multiple datasets, i.e., $\bar{\phi}$=$\frac{1}{k}\sum_{i=1}^k \phi_i$, to determine the relative impact of features across various utility functions and reveal the most influential ones. 
}
\textcolor{rebuttal}{We\todo{R3.4} use the \texttt{autorank} package~\cite{Herbold2020}, to rank the (meta-)features by their statistical significance.} 
Figures \ref{fig:cd_mfs} and \ref{fig:cd_miners} use critical difference diagrams to show feature \textcolor{minor}{\todo{R1.6}contribution}. Features connected by black bars show no statistically significant difference in rank.

\Cref{fig:cd_mfs} shows the results for six metrics across all evaluated PD algorithms. Notably, metrics \emph{Ft}, \emph{Cf}, and \emph{Sz} rank \textit{aq1} as feature with the highest \textcolor{minor}{contribution}  \textcolor{rebuttal}{\todo{M3+\\R1.4}as a low number of activities directly simplifies the process model, improving its alignment with the log while reducing complexity and size.}
Similarly, \emph{Pr} and \emph{Et} rank \textit{svo} as the most impactful feature \textcolor{rebuttal}{as a highly skewed distribution of process variants allows for the creation of simpler, more precise models that are faster to generate}.
As the harmonic mean of \emph{Ft} and \emph{Pr}, \emph{Fs} expectedly ranks \textit{aq1} and \textit{svo} as statistically equally important.
On the other end, features ranked last reflect differences between metric types.
Quality metrics (\emph{Ft}, \emph{Pr}, \emph{Fs}) and the performance metric \emph{Et} rank \textit{rt5v} are the least impactful \textcolor{rebuttal}{\todo{M3\\+R1.4}as these metrics are more influenced by the overall distribution of variants rather than the proportion of rare variants.}
Although \textit{rt5v} is similarly ranked for complexity metrics \textit{Cf}, \textit{Sz}, the lowest contribution is average by \textit{ekbr3}, \textcolor{rebuttal}{\todo{M3\\+R1.4}as a statistically unpredictable log does not always require a large or intricate model to represent it}.

\Cref{fig:cd_miners} shows the perspective per PD algorithms across all evaluated metrics. 
The features \textit{svo} and \textit{aq1} show the highest \textcolor{minor}{contribution} for all miners.
For \emph{IND} and \emph{ILP}, the rankings of \textit{svo} and \textit{aq1} show no significant difference, while \emph{SPM} ranks \textit{aq1} first and \textit{svo} second.
Likewise, \textit{rt5v} has the lowest contribution across all evaluated metrics and algorithms.
\textcolor{rebuttal}{\todo{M3+\\R1.4}
The difference for \emph{SPM} likely stems from its strategy of explicitly examining directly-follows graphs and loops, a process more influenced by the number of activities (\textit{aq1}) than variant distribution (\textit{svo}).
Conversely, \textit{rt5v} has the lowest contribution because it accounts for a small portion of the log's behavior.
More holistic features, as \textit{svo} and \textit{aq1}, are therefore on average more influential on results across all algorithms.}
\subsection{RQ2: \Paste{RQ2}}
\label{subsec:exp:featurevalues}%
\begin{figure*}[t]
\centering
\begin{subfigure}[t]{0.24\textwidth}
    \includegraphics[width=\textwidth, trim={0 0 0 0cm}, clip]{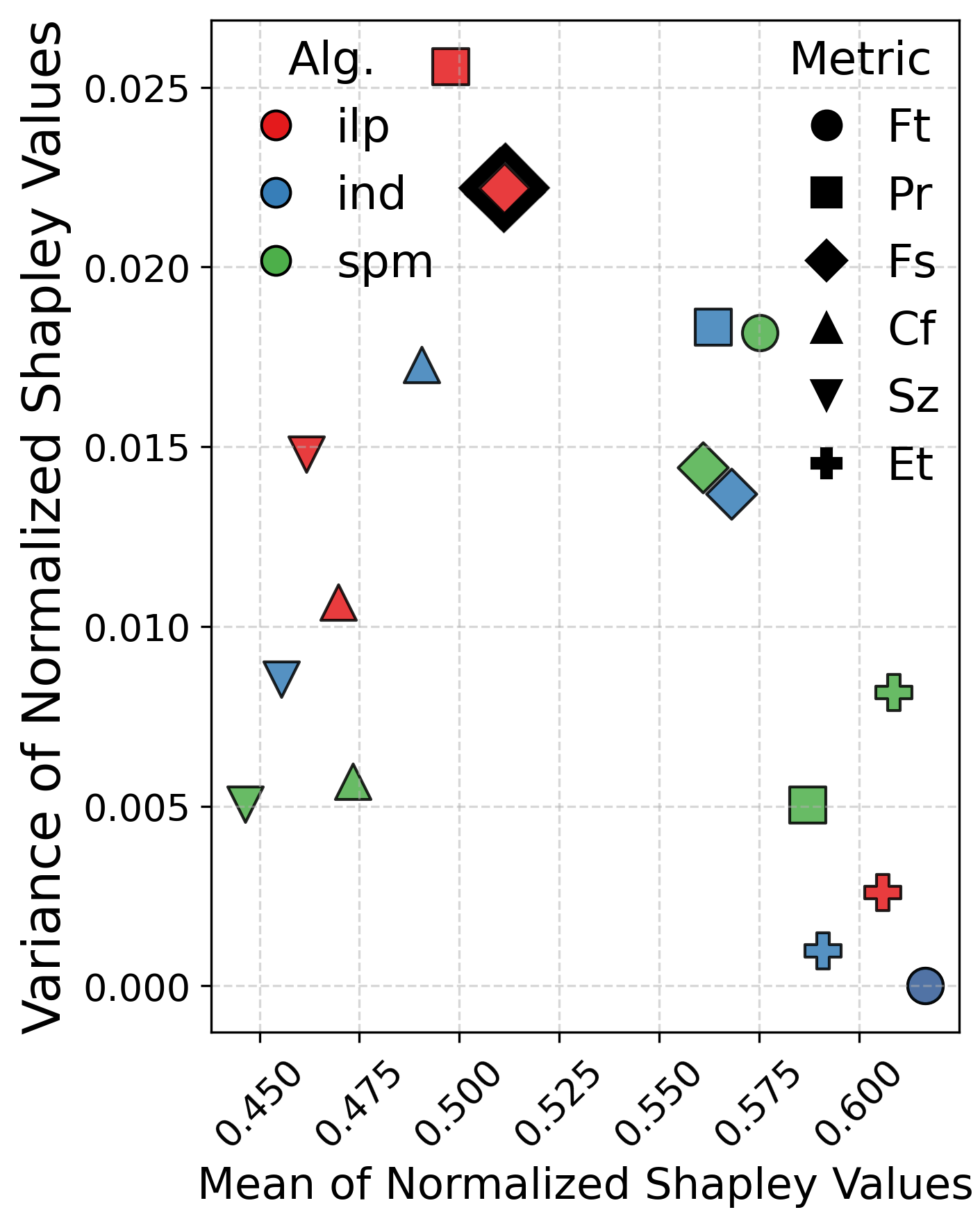}
    \caption{Robustness: Mean vs. variance}
    \label{fig:robustness}
\end{subfigure}
\hfill
\begin{subfigure}[t]{0.47\textwidth}
    \includegraphics[width=\textwidth, trim={0 0 0cm 0cm}, clip]{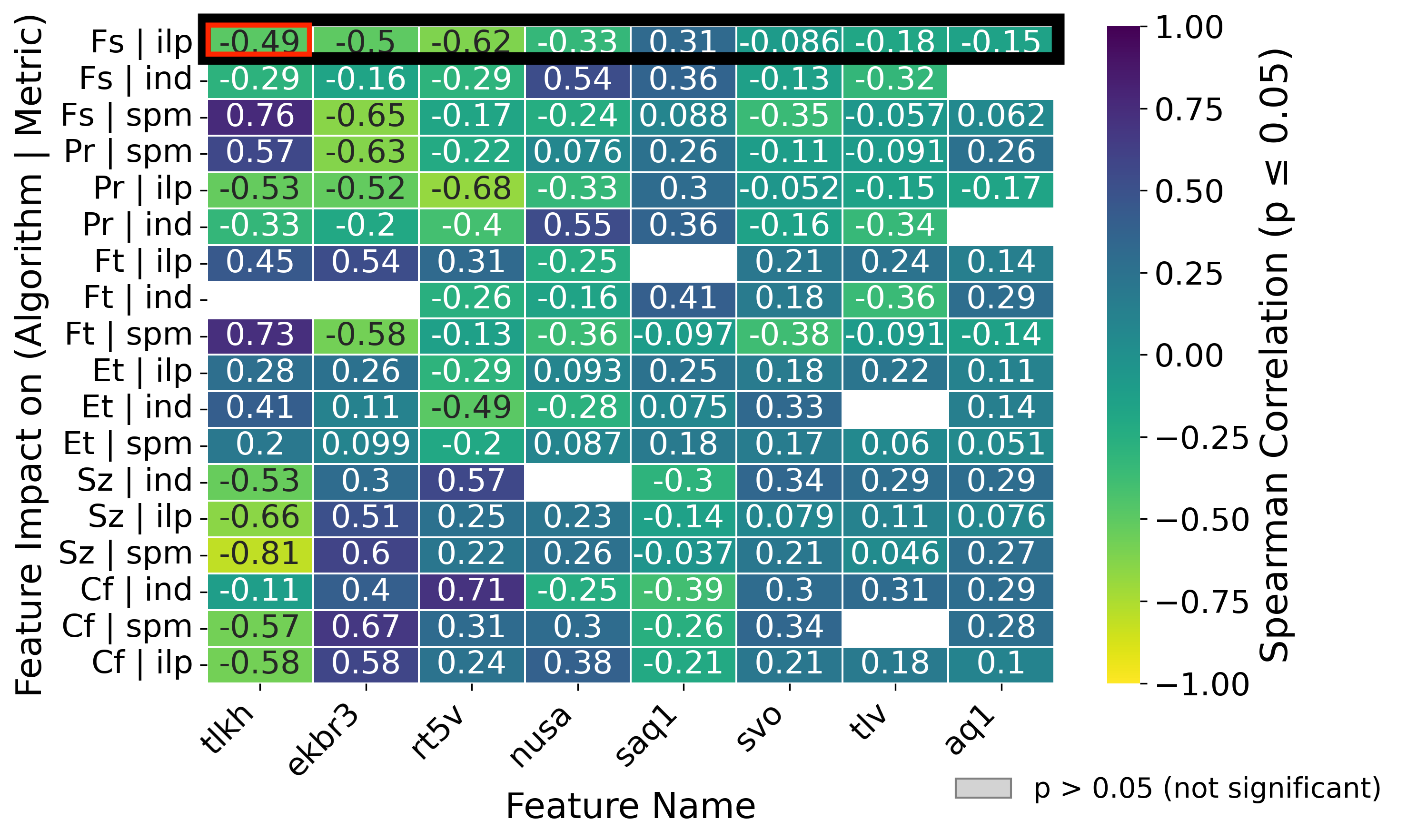}
    \caption{Correlations: between feature values and Shapley values}
    \label{fig:shap_corr}
\end{subfigure}
\hfill
\begin{subfigure}[t]{0.27\textwidth}
    \includegraphics[width=\textwidth, trim={0 0 0cm 0cm}, clip]{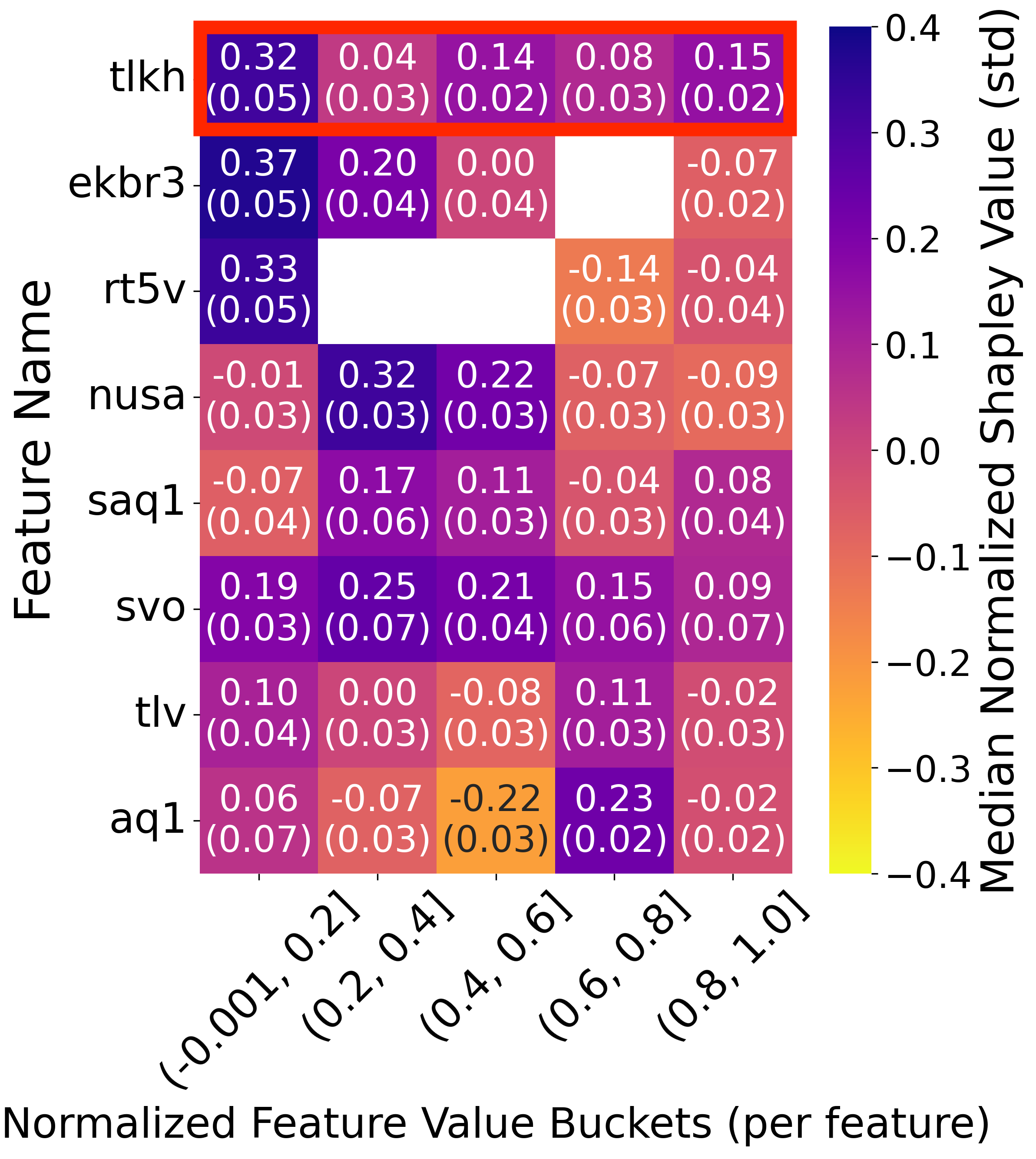}
    \caption{Feasibility: Feature values vs. Shapley values for F-score and ILP}
    \label{fig:feasibility_fscoreilp}
\end{subfigure}
\caption{\textit{Robustness}, \textit{correlation}, and \textit{feasibility} of feature values and Shapley values for process discovery algorithms}
\label{fig:value_shap}
\vspace{-1.2em}
\end{figure*}
To assess how log feature values influence process discovery results, we discuss three aspects: 
the \emph{\textcolor{minor}{robustness}\todo{R1.6}} of algorithms to feature value changes, the \emph{\textcolor{minor}{correlation}} between feature values and Shapley values, and the \emph{\textcolor{minor}{feasibility}} of producing results with a given algorithm under varying feature conditions.

\mypar{Algorithm Robustness}
To evaluate algorithm sensitivity to event log characteristics, \Cref{fig:robustness} plots the normalized mean and variance of Shapley values across features.
The mean Shapley value (\textcolor{rebuttal}{\todo{R1.3+\\R1.4}as in \emph{RQ1}}) assesses overall feature influence on each algorithm, while variance measures stability across feature combinations.
Higher mean values indicate stronger, more consistent feature influence; higher variance signals uneven responses to different feature value configurations.
All miners are affected by feature variations, but their response stability differs.
\textit{IND} shows the highest average Shapley Value for \textit{Fs}, but exhibits low variance, \textcolor{rebuttal}{\todo{M3+\\R1.4}confirming its robustness due to its structured, noise-tolerant design.}
In contrast, the \textit{ILP} is 40\% more sensitive to feature variations, despite having 7\% lower average Shapley Values. 
\textcolor{rebuttal}{\todo{M3\\+R1.4}This reflects \textit{ILP} sensitivity to infeasible or rare behavior to create precise and generalizable models.}
\textit{SPM} falls close to \textit{IND}, balancing responsiveness and stability slightly better, \textcolor{rebuttal}{likely due to its embedded filtering}.

\mypar{Correlations}
To assess how 
log feature values influence the explanatory power of features on PD outcomes, we analyze Spearman correlations between feature values and their corresponding Shapley values across algorithms and metrics. 
Spearman does not assume linearity and captures monotonic relationships.
\Cref{fig:shap_corr} shows statistically significant correlations ($p$$\leq$$0.05$) between feature values and their Shapley values, indicating whether increasing a feature consistently amplifies or reduces its impact on an evaluation metric. 
For \textit{ILP}, features like the first quartile of starting activities (\textit{saq1}) show positive correlations with quality metrics (\textit{Pr}, \textit{Fs}) but negative correlations with complexity metrics (\textit{Cf}, \textit{Sz}). 
A regular process start (high saq1) leads to better quality and simpler models,\textcolor{rebuttal}{\todo{M3+\\R1.4} as the algorithm can 
identify a few dominant entry points for the complex optimization problem
}.
In contrast, higher values of entropy for activity triplets (\textit{ekbr3}), dominance of most common variants (\textit{rt5v}), quantity of start activities (\textit{nusa}), variant occurrence skewness (\textit{svo}), and trace length variance (\textit{tlv}) negatively impact (\textit{Pr}, \textit{Fs}), while increasing (\textit{Cf}, \textit{Sz}) measurements. 
This leads to more complicated and less precise models, \textcolor{rebuttal}{\todo{M3+\\R1.4}as these features make it difficult for the miner to find a coherent solution}.
While previously discussed features have clear impact trends on metrics, the concentration of medium-length traces (\textit{tlkh}) shows a negative correlation with all discussed metrics.
I.e., if most traces are of similar length with few exceptions, \textcolor{rebuttal}{\todo{M3\\+R1.4}it can lead to overgeneralization, where \textit{ILP} produces a simpler model that fails to capture the full range of process behavior}, thereby reducing its \textit{Pr} and \textit{Fs}.

\mypar{Feasibility}
To explore the source of observed correlations and each algorithm's ability to handle different feature profiles, we use heatmaps of bucketed feature intervals versus normalized Shapley values.
\Cref{fig:feasibility_fscoreilp} shows missing \textit{Fs} values, indicating configurations where \textit{ILP} failed to produce a model. 
Feasibility issues arise for moderate-to-low \textit{rt5v} values, where moderate variability and structural ambiguity increase search complexity. In contrast, extreme \textit{rt5v} values often imply a clearer structure, making optimization more likely to succeed.
Considering occasional unsoundness and resource constraints, \autoref{tab:minersfeasibility} reports the percentage of event logs for which each algorithm produced viable models.
Despite \textit{IND}'s soundness guarantees\cite{leemans2013discovering}, its feasibility was limited: Only 39\% of logs completed within a five-minute timeout.
This reflects \textit{IND}'s vulnerability to time constraints when processing large or structurally complex logs.
\textit{SPM} achieves (76\%) feasibility using a heuristic approach that bypasses strict constraints, but its sensitivity to high variability can cause failures by prioritizing precision over model structure.
These results highlight a core challenge in process discovery: Algorithm feasibility and robustness vary with \textcolor{rebuttal}{\todo{M3}log characteristics} and computational limits.
\textit{ILP} struggles with infeasibility, \textit{IND} with timeouts, and \textit{SPM} with structural trade-offs, underscoring the need to match algorithms to log characteristics and resource constraints.%
\subsection{RQ3: \Paste{RQ3}} 
\label{subsec:exp:rq3}

\begin{table*}[t]
\begin{minipage}[t]{0.15\linewidth}
    \caption{Feasible Logs by Miner.}
    \begin{tabular}{c|c} 
         \toprule
         \makecell{\textbf{PD} \\ \textbf{Algorithm}} & \makecell{Feasible \\ Logs [\%]}\\ 
         \toprule
         IND & 39\\
         ILP & 59\\
         SPM & 76\\
         \midrule
         Overlap & 39\\
         \bottomrule
    \end{tabular}
    \label{tab:minersfeasibility}
\end{minipage}
\hfil\hfil
\begin{minipage}[t]{0.75\linewidth}
\small
\caption{Significant correlations (arrows denote direction; \textcolor{gray}{gray} if $|r| < 0.1$,\\ bold if $|r| > 0.3$, double arrow if $|r| > 0.5$)}
\label{tab:corr}
\setlength{\tabcolsep}{3pt}
\begin{tabular}{l|cccccc|cccccc|cccccc}
\toprule
\bfseries Feature & \multicolumn{6}{|c|}{\bfseries ILP} & \multicolumn{6}{|c|}{\bfseries IND} & \multicolumn{6}{|c}{\bfseries SPM} \\
\midrule 
tlkh   & \textbf{$\uparrow$Ft}    & \textbf{$\Downarrow$Pr} & \textbf{$\downarrow$Fs} & \textbf{$\Downarrow$Cf} & \textbf{$\Downarrow$Sz} & $\uparrow$Et    & 
         -      & \textbf{$\downarrow$Pr} & $\downarrow$Fs & $\downarrow$Cf & \textbf{$\Downarrow$Sz} & \textbf{$\uparrow$Et} & 
         \textbf{$\Uparrow$Ft}      & \textbf{$\Uparrow$Pr} & \textbf{$\Uparrow$Fs} & \textbf{$\Downarrow$Cf} & \textbf{$\Downarrow$Sz} & $\uparrow$Et      \\
ekbr3  & \textbf{$\Uparrow$Ft} & \textbf{$\Downarrow$Pr} & \textbf{$\Downarrow$Fs} & \textbf{$\Uparrow$Cf}    & \textbf{$\Uparrow$Sz}    & $\uparrow$Et    & 
         -      & $\downarrow$Pr & $\downarrow$Fs      & \textbf{$\uparrow$Cf}    & \textbf{$\uparrow$Sz}    & $\uparrow$Et      & 
         \textbf{$\Downarrow$Ft} & \textbf{$\Downarrow$Pr} & \textbf{$\Downarrow$Fs} & \textbf{$\Uparrow$Cf}    & \textbf{$\Uparrow$Sz}    & \textcolor{gray}{$\uparrow$Et}      \\
rt5v   & \textbf{$\uparrow$Ft}    & \textbf{$\Downarrow$Pr} & \textbf{$\Downarrow$Fs} & $\uparrow$Cf    & $\uparrow$Sz    & $\downarrow$Et & 
         $\downarrow$Ft & \textbf{$\downarrow$Pr} & $\downarrow$Fs & \textbf{$\Uparrow$Cf} & \textbf{$\Uparrow$Sz} & \textbf{$\downarrow$Et} & 
         $\downarrow$Ft & $\downarrow$Pr & $\downarrow$Fs & \textbf{$\uparrow$Cf} & $\uparrow$Sz    & $\downarrow$Et    \\
nusa   & $\downarrow$Ft & \textbf{$\downarrow$Pr}    & \textbf{$\downarrow$Fs} & \textbf{$\uparrow$Cf}    & $\uparrow$Sz      & \textcolor{gray}{$\uparrow$Et}    & 
         $\downarrow$Ft & \textbf{$\Uparrow$Pr} & \textbf{$\Uparrow$Fs}    & $\downarrow$Cf & - & $\downarrow$Et & 
         \textbf{$\downarrow$Ft}    & \textcolor{gray}{$\uparrow$Pr}    & $\downarrow$Fs & \textbf{$\uparrow$Cf} & $\uparrow$Sz    & \textcolor{gray}{$\uparrow$Et}    \\
saq1   & -      & \textbf{$\uparrow$Pr}    & \textbf{$\uparrow$Fs}    & $\downarrow$Cf    & $\downarrow$Sz & $\uparrow$Et    & 
         \textbf{$\uparrow$Ft}    & \textbf{$\uparrow$Pr}    & \textbf{$\uparrow$Fs}    & \textbf{$\downarrow$Cf}    & \textbf{$\downarrow$Sz} & \textcolor{gray}{$\uparrow$Et}      & 
         \textcolor{gray}{$\downarrow$Ft}      & $\uparrow$Pr    & \textcolor{gray}{$\uparrow$Fs}    & $\downarrow$Cf    & \textcolor{gray}{$\downarrow$Sz} & $\uparrow$Et    \\
svo    & $\uparrow$Ft    & \textcolor{gray}{$\downarrow$Pr}    & \textcolor{gray}{$\downarrow$Fs}    & $\uparrow$Cf    & \textcolor{gray}{$\uparrow$Sz}    & $\uparrow$Et    & 
         $\uparrow$Ft      & $\downarrow$Pr & $\downarrow$Fs      & \textbf{$\uparrow$Cf}    & \textbf{$\uparrow$Sz} & \textbf{$\uparrow$Et}    & 
         \textbf{$\downarrow$Ft} & $\downarrow$Pr    & \textbf{$\downarrow$Fs}    & \textbf{$\uparrow$Cf}    & $\uparrow$Sz    & $\uparrow$Et    \\
tlv    & $\uparrow$Ft    & $\downarrow$Pr      & $\downarrow$Fs    & $\uparrow$Cf    & $\uparrow$Sz    & $\uparrow$Et    & 
         \textbf{$\downarrow$Ft} & \textbf{$\downarrow$Pr} & \textbf{$\downarrow$Fs} & \textbf{$\uparrow$Cf}      & $\uparrow$Sz    & -      & 
         \textcolor{gray}{$\downarrow$Ft} & \textcolor{gray}{$\downarrow$Pr} & \textcolor{gray}{$\downarrow$Fs} & -      & \textcolor{gray}{$\uparrow$Sz}    & \textcolor{gray}{$\uparrow$Et}    \\
aq1    & $\uparrow$Ft    & $\downarrow$Pr    & $\downarrow$Fs    & $\uparrow$Cf    & \textcolor{gray}{$\uparrow$Sz}    & $\uparrow$Et    & 
         $\uparrow$Ft    & -      & -      & $\uparrow$Cf    & $\uparrow$Sz    & $\uparrow$Et    & 
         $\downarrow$Ft    & $\uparrow$Pr    & \textcolor{gray}{$\uparrow$Fs}    & $\uparrow$Cf    & $\uparrow$Sz    & \textcolor{gray}{$\uparrow$Et}      \\
\bottomrule%
\end{tabular}
\hspace*{\fill}
\end{minipage}
\vspace{-1em}
\end{table*}
\textcolor{rebuttal}{
\todo{M3\\+R1.4}
In \Cref{tab:corr} we show the correlations between feature values and their impact on results according to their Shapley values. 
A positive arrow ($\uparrow$ or $\Uparrow$) indicates that logs with larger feature values tend to co-occur with larger Shapley values.
This means that when a feature value was set higher (lower), it systematically appears as more (or less) impactful in the average marginal contribution.
However, this does not imply a causal relationship, as Shapley analysis is associational.}
We assess the utility of \ourmethod in PD from three perspectives: algorithm design, algorithm evaluation, and \textcolor{rebuttal}{\todo{M3\\+R1.4}insights to build comprehensive algorithm libraries}.

\mypar{Algorithm Design} \textit{Et} of \textit{IND} is heavily influenced by the event log's characteristics.
High \textit{Et} correlates with a greater concentration of medium-length traces (high \textit{tlkh}) and an unbalanced distribution of variant occurrences (high svo).
Conversely, logs with fewer rare variants (low r5tv) tend to reduce Et
\textcolor{rebuttal}{\todo{M3+\\R1.4}
due to \textit{IND}'s recursive, decomposing nature.}
Log characteristics that increase the depth of the recursion tree (tlkh), the number of branches (r5tv), or the difficulty of finding a suitable split (svo) increase computational complexity and thus Et.
These findings highlight \textit{IND}'s sensitivity to these features.
\textcolor{rebuttal}{\todo{M3+\\R1.4}This first invites algorithm developers to design an algorithm more robust to changes in those features regarding \textit{Et}.
Nevertheless, \textit{ILP} and \textit{SPM} are already more robust, exhibiting only an insignificant ($|r|$$<$$0.1$) to low ($0.1$$<$$|r|$$\leq0.3$) correlation between these same log characteristics and Et, thus the suggested improvement has been achieved.}
On the other hand, we find a medium ($0.3$$<$$|r|$$\leq$$0.5$) to strong ($|r|$$>$$0.5$) correlation between any other metric and at least one feature.
While trade-offs between process discovery metrics \cite{janssenswillen2017comparative} prevent one single universally robust algorithm, our analysis identifies opportunities.
\textcolor{rebuttal}{\todo{M3+\\R1.4}
These insights could guide the development of new algorithms or optimizations designed to be more robust to changes in specific log characteristics, specifically against e.g., \textit{low tlkh} and \textit{high ekbr3} for \textit{Sz}.}

\mypar{Algorithm Evaluation} \textcolor{rebuttal}{\todo{M3+\\R1.4}When evaluating a process discovery algorithm, we can leverage a feature-dropout mechanism with ablations to analyse robustness, as in our framework.
Stress tests using varying or constant event log characteristics offer transparent insight into the power and limitations of each algorithm. 
Observing the columns for Fs, between the algorithms, we can discover which feature variations they are the most robust against.}
%
Although for SPM, the concentration of medium-length traces (tlkh) and the entropy of triplets of activities (ekbr3) strongly impact the \textit{Fs} results, these are less vulnerable to variations in the quantity (nusa) and the 25th percentile of starting activities (saq1),
\textcolor{rebuttal}{\todo{M3+\\R1.4}because the algorithm's frequency-based filtering and loop/concurrency detection directly rely on the variety of whole traces (tlkh, ekbr3), but less so on single starting activities (nusa, saq1).}
In contrast, the \textit{Fs} results of \textit{IND} are most strongly and negatively impacted by a smaller set of starting activities (low nusa).
With only signs of insignificant to weak correlations for tlkh and ekbr3.
While \textit{ILP} depicts similar strongly correlated features as both \textit{IND} and \textit{SPM} for \textit{Fs}, results underline the negative impact on \textit{Fs} by high ekbr3 and a high amount of rare traces (high rt5v), \textcolor{rebuttal}{\todo{M3+\\R1.4}which introduce noise and complexity that make it difficult for the algorithm to create a precise and generalizable model.}
Likewise, a high tlkh, high nusa, or a low saq1 can also lead to overly simplified or imprecise models, further lowering \textit{Fs}.
%
\textcolor{rebuttal}{\todo{M3+\\R1.4}A novel evaluation framework could apply a masking on certain log features (e.g., collapsing of rare activity pairs, smoothing trace length variance) to test their effect on results quality. }

\mypar{Algorithm Libraries}
Taking into account the metrics' trade-offs \cite{janssenswillen2017comparative} and quantifying the impact of event log characteristics, presented in our work, process miners can build broad libraries of PD algorithms, aiming at the best results for diverse incoming event logs.
\textcolor{rebuttal}{\todo{M3+\\R1.4}For example, consider how \textit{IND} and \textit{SPM} complement each other in \textit{Fs} robustness, because their vulnerabilities are different, including disjunct strong correlation feature sets ((nusa, saq1, tlv)$\cap$(tlkh, ekbr3, svo)$=\varnothing$).
Similarly, \textit{ILP} is a good addition to the (\textit{IND}, \textit{SPM}) library, considering that it has a reduced number of features (tlkh, ekbr3, nusa), which present vulnerabilities in comparison to \textit{IND} and \textit{SPM}.
Similar arguments can be made in favor of \textit{SPM} for \textit{Ft} and \textit{Pr}.
Nevertheless, especially for \textit{Sz}, an algorithm that shows robustness (insignificant to weak correlations across all features) or vulnerability towards variations in features other than (tlkh, ekbr3), would be a great addition to this library.}

\section{Conclusion}
\label{sec:conclusion}
\textcolor{happy}{
PM algorithm evaluations often lack a systematic protocol to understand how event log characteristics influence model outcomes.
We propose SHAining, a functional Shapley analysis that assesses how meta-level configuration parameters impact the algorithm metrics by varying the input event log characteristics, instead of fixed input event logs as in prior work. 
Our method follows four steps: (i) forming feature coalitions; (ii) sampling event logs conditioned on these coalitions; (iii) applying a PM algorithm on these logs; and (iv) performing Shapley value analysis to quantify feature contribution on evaluation metrics.\\
}
\textcolor{happy}{
A large-scale evaluation across comprehensive event logs reveals how feature combinations affect different metrics in Process Discovery (PD).
We identify the most impactful features across metrics and algorithms, uncover correlations between feature values and their contributions, and provide practical insights for PD algorithm design, evaluation, and libraries.
Overall, \ourmethod enables understanding of how algorithm assumptions interact with varying event log characteristics, which is critical for PM where generalizability and model structure depend strongly on input characteristics.
}

\mypar{Threats to Validity}
While our FEEED feature selection provides broad coverage, other relevant features may exist.
To ensure more reliable analysis, we applied an intercorrelation-based feature selection procedure (cf. \Cref{sec:evaluation}) because Shapley value analysis relies on low feature intercorrelation.
Our evaluation may also be weighted towards chosen metric types and PD algorithms, which may limit the generalizability of our results by not covering all possible approaches.
Nevertheless, our modular framework allows easy incorporation of additional features, algorithms, and metrics in future work.

\mypar{Future Work} 
Future investigations should address these limitations by expanding the feature space beyond FEEED, exploring additional methods to further reduce feature intercorrelation effects on Shapley values, and incorporating a broader and more diverse set of process discovery algorithms. 
\textcolor{rebuttal}{\todo{R1.5}
We plan to explore computational optimizations, including refined sampling, to speed up our pipeline.
}
Moreover, we'll address generator bias by testing alternative generators and validating our synthetic data, as we secure diverse real-world logs.
Finally, examining industrial characteristics will further enhance our approach’s practical relevance.

\bibliographystyle{ieeetr}
\bibliography{references}

@inproceedings{lundberg2017shap,
 author = {Lundberg, Scott M and Lee, Su-In},
 booktitle = {NeurIPS},
 pages = {},
 publisher = {Curran Associates, Inc.},
 title = {A Unified Approach to Interpreting Model Predictions},
 volume = {30},
 year = {2017}
}

@inproceedings{maldonado23feeed,
  author       = {Andrea Maldonado and
                  Gabriel Marques Tavares and
                  Rafael Seidi Oyamada and
                  Paolo Ceravolo and
                  Thomas Seidl},
  title        = {{FEEED:} Feature Extraction from Event Data},
 booktitle    = {ICPM Demos},
 _publisher    = {CEUR-WS.org},
  year         = {2023},
}

@article{pishgar2022process,
  title={A process mining-deep learning approach to predict survival in a cohort of hospitalized COVID-19 patients},
  author={Pishgar, Maryam and Harford, Samuel and Theis, Julian and Galanter, William and Rodr{\'\i}guez-Fern{\'a}ndez, Jorge Mario and Chaisson, LH and Zhang, Y and Trotter, A and Kochendorfer, Karl M and others},
  journal={BMC Medical Informatics and Decision Making},
  pages={194},
  year={2022},
  publisher={Springer}
}

@inproceedings{maldonado2024gedi,
  author       = {Andrea Maldonado and
                  Christian M. M. Frey and
                  Gabriel Marques Tavares and
                  Nikolina Rehwald and
                  Thomas Seidl},
  title        = {{GEDI:} Generating Event Data with Intentional Features for Benchmarking
                  Process Mining},
  booktitle    = {BPM},
  pages        = {221--237},
  publisher    = {Springer},
  year         = {2024},
}

@inproceedings{huang2024proreco,
  title={ProReco: A Process Discovery Recommender System},
  author={Huang, Tsung-Hao and Junied, Tarek and Pegoraro, Marco and van der Aalst, Wil MP},
  booktitle={CAiSE},
  pages={93--101},
  year={2024},
  organization={Springer}
}

@inproceedings{vanden2014uncovering,
  title={Uncovering the relationship between event log characteristics and process discovery techniques},
  author={vanden Broucke, Seppe KLM and Delvaux, C{\'e}dric and Freitas, Jo{\~a}o and Rogova, Taisiia and Vanthienen, Jan and Baesens, Bart},
  booktitle={BPM Workshops},
  pages={41--53},
  year={2013},
  organization={Springer}
}

@article{augusto2019split,
  title={Split miner: automated discovery of accurate and simple business process models from event logs},
  author={Augusto, Adriano and Conforti, Raffaele and Dumas, Marlon and La Rosa, Marcello and Polyvyanyy, Artem},
  journal={KAIS},
  volume={59},
  pages={251--284},
  year={2019},
  publisher={Springer}
}

@article{augusto2018automated,
  title={Automated discovery of process models from event logs: Review and benchmark},
  author={Augusto, Adriano and Conforti, Raffaele and Dumas, Marlon and La Rosa, Marcello and Maggi, Fabrizio Maria and Marrella, Andrea and Mecella, Massimo and Soo, Allar},
  journal={IEEE TKDE},
  volume={31},
  number={4},
  pages={686--705},
  year={2018},
  publisher={IEEE}
}

@inproceedings{stevens2021quantifying,
  title={Quantifying explainability in outcome-oriented predictive process monitoring},
  author={Stevens, Alexander and De Smedt, Johannes and Peeperkorn, Jari},
  booktitle={ICPM},
  pages={194--206},
  year={2021},
  organization={Springer}
}

@article{van2018discovering,
  title={Discovering workflow nets using integer linear programming},
  author={van Zelst, Sebastiaan J and van Dongen, Boudewijn F and van der Aalst, Wil MP and Verbeek, HMW},
  journal={Computing},
  volume={100},
  pages={529--556},
  year={2018},
  publisher={Springer}
}

@article{augusto2022connection,
title = {The connection between process complexity of event sequences and models discovered by process mining},
journal = {Information Sciences},
volume = {598},
pages = {196-215},
year = {2022},
issn = {0020-0255},
_doi = {https://doi.org/10.1016/j.ins.2022.03.072},
_url = {https://www.sciencedirect.com/science/article/pii/S0020025522002997},
author = {Adriano Augusto and Jan Mendling and Maxim Vidgof and Bastian Wurm},
keywords = {Process complexity, Event sequence data, Event logs, Process mining, Graph entropy, Automated process discovery},
}

@article{adriansyah2015measuring,
  title={Measuring precision of modeled behavior},
  author={Adriansyah, Arya and Munoz-Gama, Jorge and Carmona, Josep and Van Dongen, Boudewijn F and Van Der Aalst, Wil MP},
  journal={ISeB},
  volume={13},
  pages={37--67},
  year={2015},
  publisher={Springer}
}

@inproceedings{adriansyah2011conformance,
  title={Conformance checking using cost-based fitness analysis},
  author={Adriansyah, Arya and van Dongen, Boudewijn F and van der Aalst, Wil MP},
  booktitle={EDOC},
  _pages={55--64},
  year={2011},
  _organization={IEEE}
}

@book{mendling2008metrics,
  title={Metrics for process models: empirical foundations of verification, error prediction, and guidelines for correctness},
  author={Mendling, Jan},
  _volume={6},
  year={2008},
  publisher={Springer}
}

@article{jouck2019generating,
author={Jouck, Toon
and Depaire, Beno{\^i}t},
title={Generating Artificial Data for Empirical Analysis of Control-flow Discovery Algorithms},
journal={BISE},
year={2019},
month={Dec},
day={01},
_volume={61},
_number={6},
pages={695-712},
issn={1867-0202},
_doi={10.1007/s12599-018-0541-5},
_url={https://doi.org/10.1007/s12599-018-0541-5}
}

@InProceedings{burattin2022purpose,
author="Burattin, Andrea
and Re, Barbara
and Rossi, Lorenzo
and Tiezzi, Francesco",
title="A Purpose-Guided Log Generation Framework",
booktitle="BPM",
year="2022",
publisher="Springer",
pages="181--198",
isbn="978-3-031-16103-2"
}

@inproceedings{schreiber2021exploring,
    author       = {Schreiber, Clemens},
    year         = {2021},
    title        = {Exploring the Impact of Process Diversity on Business Process Performance},
    pages        = {17-18},
    eventtitle   = {ICPM Doctoral Consortium and Demo Track},
    eventtitleaddon = {ICPM-D 2021},
    eventdate    = {2021-10-31/2021-11-04},
    venue        = {Eindhoven, Niederlande},
    booktitle    = {ICPM DC},
    publisher    = {{CEUR-WS.org}},
    issn         = {1613-0073},
    language     = {english},
    _volume       = {3098}
}

@article{6306ea88338545bc9f1781894c6736ef,
title = "Alignment-based Quality Metrics in Conformance Checking.",
author = "Dongen, {Boudewijn F. van} and Josep Carmona and Thomas Chatain",
_note = "DBLP License: DBLP's bibliographic metadata records provided through http://dblp.org/ are distributed under a Creative Commons CC0 1.0 Universal Public Domain Dedication. Although the bibliographic metadata records are provided consistent with CC0 1.0 Dedication, the content described by the metadata records is not. Content may be subject to copyright, rights of privacy, rights of publicity and other restrictions.",
year = "2016",
language = "English",
_volume = "36",
pages = "77--80",
journal = "EMISA Forum",
issn = "1610-3351",
publisher = "Gesellschaft f{\"u}r Informatik e.V. (GI)",
_number = "2",
}

@inproceedings{van1997verification,
  title={Verification of workflow nets},
  author={Van der Aalst, Wil MP},
  booktitle={Petri Nets},
  pages={407--426},
  year={1997},
  organization={Springer}
}

@inproceedings{leemans2013discovering,
  title={Discovering block-structured process models from event logs-a constructive approach},
  author={Leemans, Sander JJ and Fahland, Dirk and Van Der Aalst, Wil MP},
  booktitle={Petri Nets},
  pages={311--329},
  year={2013},
  organization={Springer}
}

@article{rehse2024process,
  title={On process discovery experimentation},
  author={Rehse, Jana-Rebecca and Leemans, S and Fettke, Peter and van der Werf, Jan Martijn E.M.},
  journal={ACM TOSEM},
  year={2024}
}

@article{janssenswillen2017comparative,
  title={A comparative study of existing quality measures for process discovery},
  author={Janssenswillen, Gert and Donders, Niels and Jouck, Toon and Depaire, Beno{\^\i}t},
  journal={Information Systems},
  volume={71},
  pages={1--15},
  year={2017},
  publisher={Elsevier}
}

@inproceedings{van2021all,
  title={All that glitters is not gold: Towards process discovery techniques with guarantees},
  author={van der Werf, Jan Martijn EM and Polyvyanyy, Artem and van Wensveen, Bart R and Brinkhuis, Matthieu and Reijers, Hajo A},
  booktitle={CAiSE},
  pages={141--157},
  year={2021},
  organization={Springer}
}

@article{6138a20588854e0182f4aa3595788ac6,
title = "Entropy as a Measure of Log Variability",
abstract = "Process mining algorithms fall in two classes: imperative miners output flow diagrams, showing all possible paths, whereas declarative miners output constraints, showing the rules governing a process. But given a log, how do we know which of the two to apply? Assuming that logs exhibiting a large degree of variability are more suited for declarative miners, we can attempt to answer this question by defining a suitable measure of the variability of the log. This paper reports on an exploratory study into the use of entropy measures as metrics of variability. We survey notions of entropy used, e.g. in physics; we propose variant notions likely more suitable for the field of process mining; we provide an implementation of every entropy notion discussed; and we report entropy measures for a collection of both synthetic and real-life logs. Finally, based on anecdotal indications of which logs are better suited for declarative/imperative mining, we identify the most promising measures for future studies. For estimating overall entropy, global block and k-nearest neighbour estimators of entropy appear most promising and excel at identifying noise in logs. For estimating entropy rate we identify Lempel–Ziv and certain variants of k-block estimators performing well, and note that the former is more stable, but sensitive to noise, while the latter is less stable, being sensitive to cut-off constraints determining block size. ",
keywords = "Faculty of Science, Process Mining, Hybrid Models, Process Variability, Process Flexbility, Information Theory, Entropy, Knowledge Work",
author = "Back, {Christoffer Olling} and S{\o}ren Debois and Tijs Slaats",
year = "2019",
month = jun,
day = "14",
_doi = "10.1007/s13740-019-00105-3",
language = "English",
volume = "8",
pages = "129–156",
journal = "Journal on Data Semantics",
issn = "1861-2032",
publisher = "springer verlag (germany)",
number = "2",
}

@article{munoz2022process,
  title={Process mining for healthcare: Characteristics and challenges},
  author={Munoz-Gama, Jorge and Martin, Niels and Fernandez-Llatas, Carlos and Johnson, Owen A and Sep{\'u}lveda, Marcos and Helm, Emmanuel and Galvez-Yanjari, Victor and Rojas, Eric and Martinez-Millana, Antonio and others},
  journal={JBI},
  volume={127},
  pages={103994},
  year={2022},
  publisher={Elsevier}
}

@inproceedings{carmona2017summary,
  title={Summary of the process discovery contest 2016},
  author={Carmona, Josep and de Leoni, Massimiliano and Depaire, Beno{\^\i}t and Jouck, Toon},
  year={2017},
  organization={Springer},
booktitle={~}
}

@InProceedings{andree2024workflow,
author="Andree, Kerstin
and Hoang, Mai
and Dannenberg, Felix
and Weber, Ingo
and Pufahl, Luise",
title="Discovery of Workflow Patterns - A Comparison of Process Discovery Algorithms",
booktitle="CoopIS",
year="2024",
publisher="Springer", 
_publisher="Nature Switzerland",
_address="Cham",
pages="257--274",
isbn="978-3-031-46846-9"
}

@article{Lipovetsky_2020_handbook,
author = {Stan Lipovetsky},
title = {Handbook of the Shapley Value},
journal = {Technometrics},
volume = {62},
number = {2},
pages = {1--280},
year = {2020},
publisher = {ASA Website},
doi = {10.1080/00401706.2020.1744904},
_URL = {https://doi.org/10.1080/00401706.2020.1744904},
_eprint = {https://doi.org/10.1080/00401706.2020.1744904}
}

@book{1952_shapley,
author="Shapley, Lloyd S.",
title="A Value for N-Person Games",
_address="Santa Monica, CA",
year="1952",
doi="10.7249/P0295",
publisher="RAND Corporation"
}

@inproceedings{heskes2020causal,
  title={Causal Shapley Values: Exploiting Causal Knowledge to Explain Individual Predictions of Complex Models},
  author={Heskes, Tom and Sijben, Emiel and Bucur, Ioana G and Claassen, Tom},
  booktitle={NeurIPS},
  year={2020},
  _note={arXiv:2011.01625},
  url={https://arxiv.org/abs/2011.01625}
}

@article{Herbold2020,
  _doi = {10.21105/joss.02173},
  _url = {https://doi.org/10.21105/joss.02173},
  year = {2020},
  publisher = {The Open Journal},
  _volume = {5},
  _number = {48},
  pages = {2173},
  author = {Steffen Herbold},
  title = {Autorank: A Python package for automated ranking of classifiers},
  journal = {Journal of Open Source Software}
}

@article{explainabilityNotAGame,
author = {Jo{\~{a}}o Marques{-}Silva and
Xuanxiang Huang},
title = {Explainability Is \emph{Not} a Game},
journal = {Commun. {ACM}},
volume = {67},
number = {7},
pages = {66--75},
year = {2024},
_url = {https://doi.org/10.1145/3635301},
doi = {10.1145/3635301},
timestamp = {Mon, 05 Aug 2024 09:54:20 +0200},
_biburl = {https://dblp.org/rec/journals/cacm/MarquesSilvaH24.bib},
bibsource = {dblp computer science bibliography, https://dblp.org}
}

@inproceedings{2024_logicBasedXAI,
author = {Marques-Silva, Joao},
title = {Logic-Based Explainability: Past, Present and Future},
year = {2024},
isbn = {978-3-031-75386-2},
publisher = {Springer-Verlag},
_address = {Berlin, Heidelberg},
_url = {https://doi.org/10.1007/978-3-031-75387-9_12},
_doi = {10.1007/978-3-031-75387-9_12},
_booktitle = {Leveraging Applications of Formal Methods, Verification and Validation. Software Engineering Methodologies: 12th International Symposium, ISoLA 2024, Crete, Greece, October 27–31, 2024, Proceedings, Part IV},
booktitle = {12th International Symposium, ISoLA 2024, October 27–31, 2024, Proceedings, Part IV},
pages = {181–204},
numpages = {24},
keywords = {Explainable AI, Symbolic AI, Formal Explainability, Certification},
location = {Crete, Greece}
}

@InProceedings{kwon_dshapley_2021,
  title = 	 { Efficient Computation and Analysis of Distributional Shapley Values },
  author =       {Kwon, Yongchan and A. Rivas, Manuel and Zou, James},
  booktitle = 	 {Proceedings of The 24th International Conference on Artificial Intelligence and Statistics},
  pages = 	 {793--801},
  year = 	 {2021},
  _editor = 	 {Banerjee, Arindam and Fukumizu, Kenji},
  _volume = 	 {130},
  _series = 	 {Proceedings of Machine Learning Research},
  _month = 	 {13--15 Apr},
  _publisher =    {PMLR},
  _url = 	 {https://proceedings.mlr.press/v130/kwon21a.html},
}

\end{document}